\pdfoutput=1

\documentclass[11pt]{article}

\usepackage{acl}

\usepackage{times}
\usepackage{latexsym}

\usepackage[T1]{fontenc}

\usepackage[utf8]{inputenc}

\usepackage{microtype}

\usepackage{multirow,booktabs,makecell}
\usepackage{booktabs}
\usepackage{graphicx}
\usepackage{adjustbox}
\usepackage{color}
\usepackage{xspace}
\usepackage{enumitem}

\title{End-to-end Dense Video Captioning as Sequence Generation}

\author{
Wanrong Zhu\textsuperscript{1}, 
Bo Pang\textsuperscript{2},
Ashish V. Thapliyal\textsuperscript{2}, 
William Yang Wang\textsuperscript{1}, 
Radu Soricut\textsuperscript{2} \\
\textsuperscript{1}UC Santa Barbara, 
\textsuperscript{2}Google Research\\
\texttt{\small \{wanrongzhu,william\}@cs.ucsb.edu}
\texttt{\small \{bopang,asht,rsoricut\}@google.com}
}

\begin{document}
\maketitle

\newcommand{\TODO}[1]{\textcolor{red}{[TODO]: #1}}
\newcommand{\partitiontask}[0]{modified dense video captioning\xspace}
\newcommand{\partitionsetting}[0]{modified setting\xspace}
\newcommand{\Partitiontask}[0]{Modified dense video captioning\xspace}
\newcommand{\Vanilla}{Original\xspace}
\newcommand{\vanilla}{original\xspace}
\newcommand{\Hardencoding}[0]{Tagging-based\xspace}
\newcommand{\Offsetbased}[0]{Length-based\xspace}
\newcommand{\hardencoding}[0]{tagging-based\xspace}
\newcommand{\offsetbased}[0]{length-based\xspace}
\newcommand{\timestampbased}[0]{Timestamp markers\xspace}
\newcommand{\TemporalEmb}{Temporal embedding\xspace}
\newcommand{\temporalEmb}{temporal embedding\xspace}
\newcommand{\TemporalEmbShort}{Emb\textsubscript{\sc{Time}}\xspace}
\newcommand{\SimpleConcat}{Simple Concatenation\xspace}
\newcommand{\SimpleConcatShort}{SimpleConcat\xspace}
\newcommand{\notemporal}{\SimpleConcatShort}
\newcommand{\timeanchorShort}{T-marker\xspace}
\newcommand{\withtemporal}{+ \TemporalEmbShort}
\newcommand{\withanchor}{+ \timeanchorShort}
\newcommand{\segonly}{Seg-only model\xspace}
\newcommand{\segcap}{Seg+Cap model\xspace}
\newcommand{\sep}{$\langle$sep$\rangle$}
\newcommand{\pad}{$\langle$pad$\rangle$}


\begin{abstract}
Dense video captioning aims to identify the events of interest in an input video, and generate descriptive captions for each event.
Previous approaches usually follow a two-stage generative process, which first proposes a segment for each event, then renders a caption for each identified segment.
Recent advances in large-scale sequence generation pretraining have seen great success in unifying task formulation for a great variety of tasks, but so far, more complex tasks such as dense video captioning are not able to fully utilize this powerful paradigm.
In this work, we show how to model the two subtasks of dense video captioning jointly as {\em one} sequence generation task, and simultaneously predict the events and the corresponding descriptions.
Experiments on YouCook2 and ViTT show encouraging results and indicate the feasibility of training complex tasks such as end-to-end dense video captioning integrated into large-scale pretrained models.
\end{abstract}
\section{Introduction}

Online videos have become an important source of knowledge and skills~\citep{oneil-hart_2017}. In order to help users locate information of interest, search engines and video platforms often show anchors at ``key moments'', usually accompanied by descriptions of the segment's content~\citep{baheti_2019}.
This is a direct application of the dense video captioning task~\citep{Krishna2017DenseCaptioningEI}, thus methods for improving performance on this task are  relevant to any video platform.

Intuitively, dense video captioning can be decomposed into two subtasks: event localization and segment-level video captioning. Following this approach, 
prior work~\citep{Krishna2017DenseCaptioningEI,ZhXuCoAAAI18,Li2018JointlyLA,Wang2018BidirectionalAF,Zhou2018EndtoEndDV,Mun2019StreamlinedDV,Iashin2020MultimodalDV} has used a two-stage, ``localize-then-describe'' pipeline.
Such methods usually involve two separate modules with different underlying model architectures for event localization and event captioning, with captions for dense events rendered based on the predicted event spans. 

Recently, with the advance of large-scale datasets and model architectures, there has been an explosion of pretrained multimodal (for text, image, video) Transformer models~\citep{Tan2019LXMERTLC,Sun2019VideoBERTAJ, Li2019VisualBERTAS, Luo2020UniViLMAU, Li2020UnicoderVLAU,Li2020OscarOA, Gan2020LargeScaleAT, Kim2021ViLTVT}.
Such models have proved to be highly effective when fine-tuned for a wide range of downstream tasks, such as visual question answering~\citep{Agrawal2015VQAVQ}, image captioning~\citep{cococap}, visual common sense reasoning~\citep{Zellers2019FromRT}, visual entailment~\citep{Xie2019VisualEA}, etc. 
These end-tasks can be expressed as sequence generation tasks in a straightforward manner.
In contrast, this is non-trivial for dense video captioning, as the segmentation subtask does not lend itself naturally to such a formulation.
Does this mean more complex tasks cannot benefit from the pretraining paradigm in an end-to-end fashion?
In this work, we study dense video captioning as an example of a complex task that can be cast as sequence generation and, as a result, can benefit from large-scale pretraining.

More specifically, we propose to solve the dense video captioning task as a single sequence-to-sequence modeling task using a multimodal Transformer.
To this end, we design several task formulations to encode both segmentation and caption prediction in one target string. Thus, our task formulations allow the model to simultaneously predict event locations and corresponding captions in one pass, using one decoder. 
This opens the door to leveraging large-scale pretrained models, as well as the option of participating in large-scale multi-task training more easily by reusing existing infrastructure.

We evaluate our model on two dense video captioning benchmarks, YouCook2~\citep{ZhXuCoAAAI18} and ViTT~\citep{Huang2020MultimodalPF}.  Our sequence generation formulations provide a feasible path forward -- we obtain encouraging results compared to prior work that used a two-stage scheme with specialized architectures for each step.  On the pretraining front: (a) we are able to benefit from models pretrained on very different data and tasks, such as T5~\cite{t5}, (b) pretraining on more domain-specific data (WikiHow) and pretraining tasks (predicting headings for how-to steps) lead to a similar amount of gain, but (c) having the domain-specific pretraining start from a T5 checkpoint (T5 + WikiHow) provides a significantly larger gain.  
The noteworthy result is that, even in the presence of large-scale domain- and task-specific pretraining (WikiHow), one can still observe measurable benefits from a task-agnostic general-purpose pretrained model (T5).

While the primary motivation for modeling the two tasks jointly is to be able to utilize the pretraining paradigm, the segmentation subtask (finding event boundaries) and the captioning subtask (describing what happens in an event) are related tasks, and intuitively stand to benefit from being modeled jointly.  Our experimental results are aligned with this intuition: a model that does both segmentation and captioning simultaneously, outperforms (in terms of segmentation accuracy) a variant that focuses only on the segmentation task.

Overall, our results point to a viable alternative direction for modeling complex tasks such as end-to-end dense video captioning, in which we can leverage the large-scale pretraining paradigm to achieve modeling improvements.
\section{Related Work}

\subsection{Multimodal Transformer}
Recently, vision-and-language pre-training has attracted a lot of attention for jointly learning from visual and textual inputs in order to better solve multimodal tasks. 
Following the success of BERT~\citep{Devlin2019BERTPO}, multimodal pre-training usually adopts the Transformer~\citep{Vaswani2017AttentionIA} encoder structure to encode both the visual features and textual features.
The late-fusion approaches first process visual and textual information separately and subsequently fuse them using another Transformer layer~\citep{Tan2019LXMERTLC, Lu2019ViLBERTPT}. 
The early-fusion approaches jointly encode visual and texual representations~\citep{Chen2020UNITERUI, Sun2019VideoBERTAJ, Li2019VisualBERTAS, Luo2020UniViLMAU, Li2020UnicoderVLAU, Qi2020ImageBERTCP, Huang2020PixelBERTAI, Li2020OscarOA, Lin2020InterBERTVI, Gan2020LargeScaleAT, Kim2021ViLTVT}.
During pre-training, tasks such as masked language modeling, masked region modeling, and image-text matching are used to learn a cross-modal encoding which benefits downstream multimodal tasks. 

\subsection{Dense Video Captioning}
 \citet{Krishna2017DenseCaptioningEI} introduced the dense video captioning (DVC) task and proposed a solution based on two separate modules: one for proposing events, and another for captioning them. 
Recent work~\citep{ZhXuCoAAAI18,Li2018JointlyLA,Wang2018BidirectionalAF,Zhou2018EndtoEndDV,Mun2019StreamlinedDV,Iashin2020MultimodalDV} follows the two-stage ``detect-then-describe'' framework, in which the event proposal module first predicts a set of event segments, then the captioning module constructs captions for each candidate event segment.
Another line of work~\citep{Deng2021SketchGA,Wang2021EndtoEndDV} removes the explicit event proposing process. \citet{Deng2021SketchGA} tackles the DVC task from a top-down perspective, in which they first generate a video-level story, then ground each sentence in the story into a video segment.
\citet{Wang2021EndtoEndDV} considers the DVC task as a set prediction problem, and applies two parallel prediction heads for event localization and captioning. 
To the best of our knowledge, our work is the first to simultaneously conduct event localization and captioning in a single run\footnote{
Note that on a different task (object detection), contemporaneous work~\citep{chen2022pixseq} has combined spatial localization and object description via a sequence generation formulation by predicting bounding box coordinates and object labels in sequence.} within the same prediction head for the dense video captioning task.


\section{Task Definition}
\label{sec:task_definition}

\begin{figure*}[tbp]
    \centering
    \resizebox{\linewidth}{!}{%
    \includegraphics{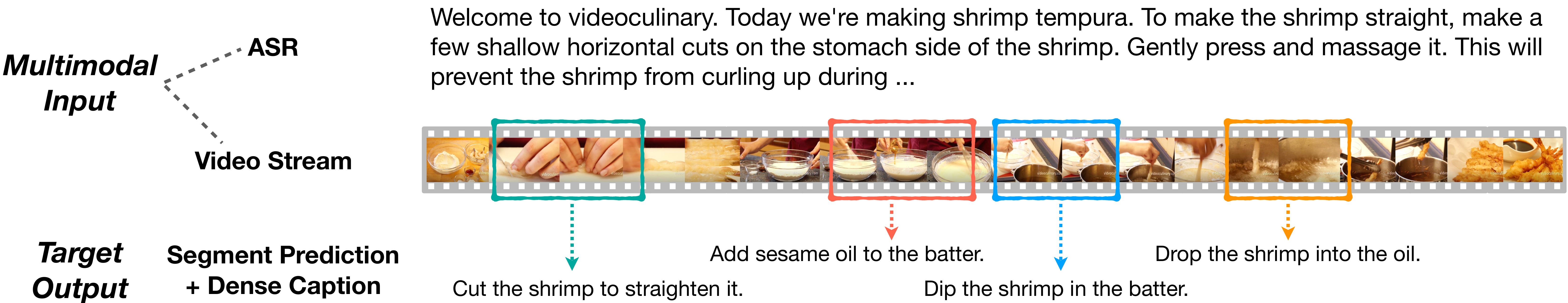}
    }
    \caption{An example of the input video and output segmentations and captions from the YouCook2 dataset.
}
    \label{fig:youcook2_eg}
\end{figure*}

\begin{figure}[tbp]
    \centering
    \includegraphics[width=\linewidth]{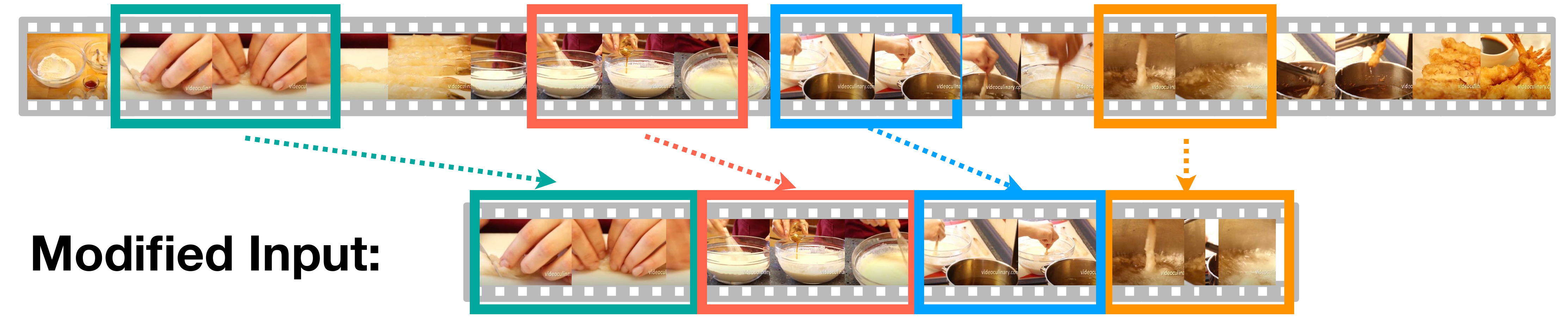}
    \caption{\Partitiontask: a simplified setting where the segments are concatenated to form the modified input with gaps removed. 
    Table~\ref{tab:youcook2_partition},~\ref{tab:ablation_input_source},~\ref{tab:ablation_t5_ckpt} show our preliminary experiments using this set up; all other results reported in the paper are carried out in the original setting as depicted in Figure~\ref{fig:youcook2_eg}.
    }
    \label{fig:partition_eg}
\end{figure}

The DVC task consists of annotating each input video into multiple segments, where each segment corresponds to an event of interest accompanied by a short description (caption).  
Figure~\ref{fig:youcook2_eg} shows an example from the YouCook2 dataset.

\paragraph{\Partitiontask}
In YouCook2, each segment is marked by a start and an end time, often with gaps between segments. 
The burden of identifying not just the right start-time but also the right end-time increases the difficulty of the segmentation task. 
Thus, we start our exploration with a simpler task where we introduce a variant of the YouCook2 dataset as shown in Fig.~\ref{fig:partition_eg}: all the annotated segments in a given video are concatenated to form a {\em modified} input, leaving out the gaps between segments. 
We refer to this setting as the {\em \partitiontask}:
given an input from Fig.~\ref{fig:partition_eg}, the model only needs to predict $n$ start times to fully define $n$ segments.
In this setting, the segmentation subtask becomes a {\em partition} task for identifying the set of start times of segments.

\section{Method}
\label{sec:model}

\begin{figure*}[htbp]
    \centering
    \vspace{-1ex}
    \includegraphics[width=\linewidth]{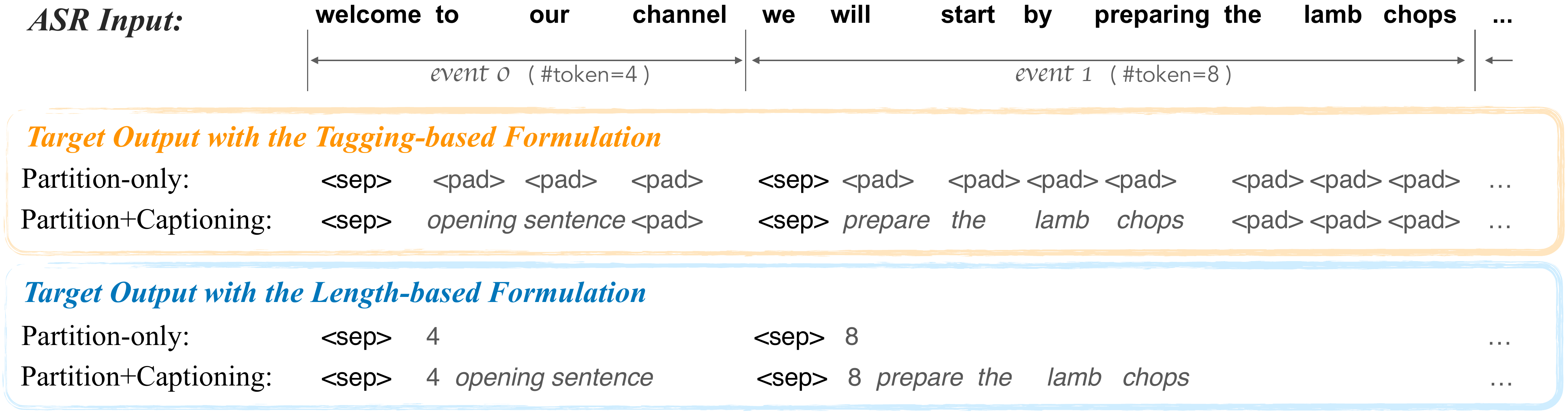}
    \caption{The \hardencoding and \offsetbased target formulations for \partitiontask.
    }
    \label{fig:formulations}
\vspace{-1ex}
\end{figure*}

As noted earlier, prior work often decomposes dense video captioning into two subtasks, (a) a segmentation subtask, and (b) a segment-level captioning subtask.
These two subtasks are often addressed with different model architectures.
In contrast, our approach solves both subtasks simultaneously with one single model.  

We first describe how we jointly model segmentation and captioning subtasks as one single sequence generation task.  To this end, we need to formulate target strings in ways that encode both segmentation and captioning predictions.

The typical input to a DVC task includes both visual information and speech in textual form -- Automatic Speech Recognition (ASR) tokens.
We start by introducing our target string formulations assuming only textual input, with segmentation information expressed in terms of the positions of the corresponding ASR tokens\footnote{Our motivation for treating DVC as a sequence generation task is to take advantage of existing pretrained sequence generation models, currently dominated by text models; thus, we take a text-centric view in this work.}.  We then describe multi-modal models where the visual information is added to the input while retaining the aforementioned scheme to represent segmentation information.

\subsection{Target string formulations} 
We describe two approaches to formulate the target strings.  We refer to a model that encodes only segmentation information in the target strings as a \textbf{\segonly}, and one that encodes both segmentation and captioning as a \textbf{\segcap}.

\paragraph{\Hardencoding target formulation}
We encode the segmentation subtask in a manner similar to the encoding of the chunking task as tagging tokens in the IOB format \cite{ramshaw-marcus-1995-text}.  
Fig.~\ref{fig:formulations} illustrates how we model the segmentation task with two tags (in the \partitionsetting):
the ASR token at the start of a segment receives a special token \sep{} as the start-of-segment tag, and the rest of the tokens in the segment receive a continuation tag (we reuse the \pad{} token).  
This can be extended to cover the \vanilla setting (with gaps between segments) with an additional end-of-segment tag.  
In this formulation, the ground-truth target output string has the exact same length as the input ASR string.
To model the captioning annotation, the \sep{} token is followed by the corresponding ground-truth caption, which is then padded till the next \sep{} token. 

While treating the segmentation task as a tagging task seems natural,
the \hardencoding formulation enforces equal lengths between predicted output and the input ASR tokens, which leads to potential inefficiencies: 
the input ASR string is usually much longer than all the descriptive captions combined, which results in many padding tokens in the target output, and leads to an unnecessary slow-down in training and prediction time.   
Additionally, the longer-form target strings are markedly different from the usual generative pattern of the pretrained text decoder, which can reduce the effectiveness of the pretrained checkpoints.
Furthermore, this formulation also assumes that captions are shorter than the ASR string for each segment; while this is mostly true, for segments where little is being explained (short ASR string), this formulation leaves insufficient capacity in the target string between the two consecutive \sep{} tags to encode the corresponding caption, resulting in caption truncation.

\paragraph{\Offsetbased target formulation}
To cope with the limitations of the \hardencoding formulation, we predict the {\em length} of each segment explicitly.

Let $l_i$ be the number of ASR tokens in the $i$-th segment.
In the \partitionsetting, the segmentation information for an input string with $n$ segments is fully specified by the sequence $\{l_1, l_2, ..., l_n\}$. 
Fig.~\ref{fig:formulations} provides an example of this \offsetbased formulation.
The ground-truth target string in a \segonly is simply a sequence of numbers corresponding to segment lengths (measured by the number of tokens); in a \segcap, each number is followed by the caption for that segment.

In the \vanilla setting with gaps between segments,
let $g_i$ be the offset from the last ASR token in the previous segment to the start of segment $i$.  
The target string will now aim to predict both ($g_i$, $l_i$) instead of just $l_i$ for each segment.
The sequence of all ($g_i$, $l_i$) will fully specify all segment boundaries and can be used to compute the index of the start and end ASR tokens for each segment.  

This formulation has the advantage of a more efficient representation of the segmentation information, and thus a much shorter target length.
The segmentation information is now explicitly expressed as numbers in the target strings, so the model needs to both figure out segmentation boundaries {\em and} be able to count appropriately.
We explicitly want to empirically measure the ability of our models to do the latter.

\subsection{Input formulation for multimodal signals}

\paragraph{\SimpleConcat (\SimpleConcatShort)}
Visual information for a given video is represented as a fixed-length sequence of pre-computed frame-level features.
These features are projected to the token embedding space via a fully connected layer.
We simply concatenate the sequence of ASR token embeddings and the sequence of projected visual features to form the multimodal input to the encoder.  There's one potential caveat: while the visual features are extracted at a fixed frame rate, the ASR tokens are often {\em not} spoken at a fixed speed; thus positions in this multimodal input sequence do not provide straightforward information on which visual frames are temporally aligned with a certain ASR span.  Since segmentation prediction is expressed relative to the ASR-token position index, it is not clear whether the model is able to take full advantage of visual information, absent how these two modalities align temporally.

Prior work on multimodal pretraining has found visual-textual information alignment to be a reasonably solvable task.
\newcite{Huang2020MultimodalPF} reported 87\% accuracy for aligning video segments and ASR spans in HowTo100M \cite{miech2019howto100m}, so it is possible that the decoder can learn to attend to appropriate visual information while ``counting'' the ASR tokens.

\paragraph{\TemporalEmb (\TemporalEmbShort)}
We can also express the temporal alignment more explicitly in the input by adding temporal embeddings to both ASR tokens and visual frames. 
In this formulation, we learn a temporal embedder shared between the text modality and the visual modality, which maps timestamps to temporal embeddings.  Embeddings computed from token timestamps are then added to ASR token embeddings, and embeddings computed from frame timestamps are added to projected visual frame features.  This way, ASR tokens and frames that are temporally close to each other receive similar temporal embeddings, making their representations closer to each other.


For more explorations on explicitly expressing temporal alignment in the input, see Appendix \ref{sec:appendix:markers} for an additional method to insert explicit timestamp markers into the input text sequence.

\begin{table*}[tbp]
\begin{adjustbox}{width=\linewidth,center}
\begin{tabular}{l | l | l | r r r r | r r r r r r}
\cmidrule[\heavyrulewidth]{1-13}
\multirow{2}{*}{\textbf{*}}    & \multirowcell{2}{\textbf{Target}\\\textbf{Formulation}}    & \multirow{2}{*}{\textbf{Checkpoint}}  &    \multicolumn{4}{c}{\textbf{\segonly}}  &   \multicolumn{6}{|c}{\textbf{\segcap}} \\
\cmidrule{4-13}
    &   &    &  mIoU & Precision  & Recall    &   F1  &  mIoU &   F1  &   B@4    &   METEOR  &   CIDEr   &   ROUGE-L \\ 
\cmidrule[\heavyrulewidth]{1-13}

\texttt{0} & \multicolumn{2}{c|}{Random Partition}    & 	37.74 & 	26.13 & 	24.88 & 	23.52   &   -   &   -   &   -      &   -      &   -      &   -   \\
\cmidrule{1-13}

\texttt{1} &\multirow{2}{*}{\Hardencoding}
    &   -        &	33.59	&	23.04	&	29.37	&	24.46	       &	19.72	&	17.09	&	0.07	&	0.91	&	0.03	&	2.07	  \\
\texttt{2} &    &   T5     &	12.06	&	1.78	&	7.46	&	2.81         &	6.73	&	0.24	&	0.00	&	0.01	&	0.00	&	0.03	\\
\cmidrule{1-13}

\texttt{3} &\multirow{2}{*}{\Offsetbased}
    &   -        &	36.30	&	26.23	&	28.79	&	25.81	       &	33.62	&	24.69	&	0.24	&	1.62	&	0.04	&	4.03	\\
\texttt{4} &    &   T5      &	42.71	&	31.85	&	33.04	&	31.21        &	42.82	&	32.16	&	1.83	&	4.17	&	0.21	&	8.74	\\

\cmidrule[\heavyrulewidth]{1-13}
\end{tabular}
\end{adjustbox}
\caption{Preliminary experiments comparing the \hardencoding and the \offsetbased formulation on YouCook2 \partitiontask. We report the evaluation results on the validation set (one run per setting) with models initialized from random weights or from T5 checkpoints.   }
\label{tab:youcook2_partition}
\end{table*}

\section{Experiments}
\subsection{Datasets}
\paragraph{Dense Video Captioning Datasets}
We use two publicly available datasets to verify the effectiveness of our model formulations: YouCook2~\citep{ZhXuCoAAAI18} and ViTT~\citep{Huang2020MultimodalPF}.\footnote{YouCook2 released under an MIT license; ViTT released under an ``AS IS'' license.}
The YouCook2 dataset is restricted to videos retrieved from YouTube from the cooking domain, targeting 89 recipes; each event segment is manually marked with a start and end time, along with a human-generated caption for each tightly-bounded segment.
The ViTT dataset contains instructional videos from YouTube-8M~\citep{AbuElHaija2016YouTube8MAL} and covers a broader range of topics. Its segment annotation focuses on event start time and rater-provided captions for the corresponding segment (spanning two consecutive start-time annotations).
Both datasets are annotated with captions written in English.  

Note that while the YouCook2 data release contains training, dev, and test sets, its test set does not come with human annotations. 
Thus, we split the original validation set into validation and test splits for our experiments.
For ViTT, we use the original train/val/test splits provided with the data.
The number of videos available for use at the time of our work\footnote{As of 2021; note that YouTube videos are subject to user deletion.} for Youcook2 is 925 for train, 206 for validation and 105 for test. For ViTT there are 4736 train, 932 validation and 932 test videos.

\paragraph{Domain-specific pretraining with WikiHow}
In addition to general-purpose pre-trained models like T5, we also experiment with domain-specific pretraining.
To this end, we use the WikiHow dataset~\citep{Koupaee2018WikiHowAL}. 
WikiHow consists of instructional (how-to) articles, which makes it {\em in-domain data} for the two dense video captioning datasets considered here, while being much larger in size\footnote{WikiHow has 157,116 articles in its training set, and 5,593 articles in its validation set.}.
In addition, WikiHow articles contain detailed step-by-step instructions. Each step comes with a summary, which usually serves as the section title.
Both the step boundaries and summaries are easily extracted according to the page meta-data.
This provides the ground-truth annotation for a ``dense document caption'' task: given the full article as a sequence of text tokens,
predict the step boundaries and summaries.
This enables us to also include a {\em domain-specific pretraining task} that closely resembles our task.
For each formulation described in Sec.~\ref{sec:model}, we experiment with a checkpoint pre-trained on the WikiHow data using the corresponding target string formulation.

\subsection{Evaluation Metrics}

\begin{figure}[tbp]
    \centering
    \includegraphics[width=\linewidth]{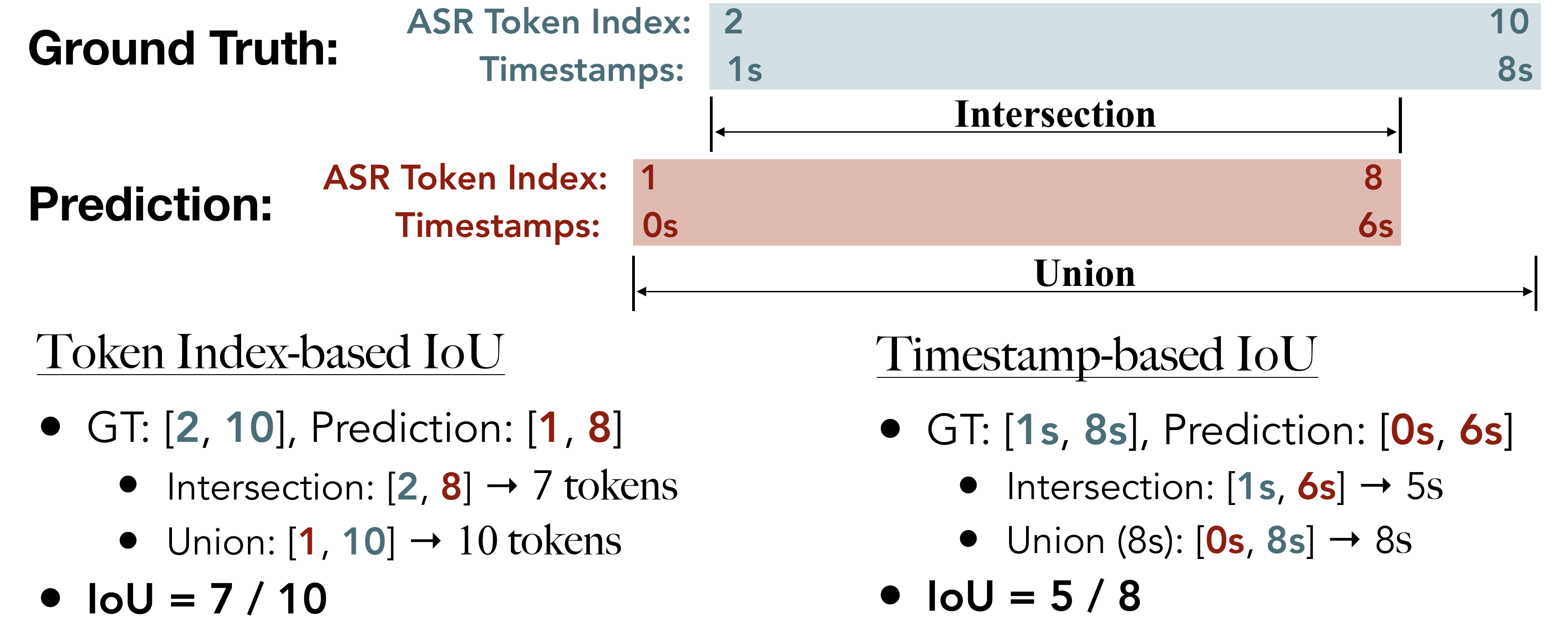}
    \caption{Comparisons between the token index-based and timestamp-based IoU used in our study.
    }
    \label{fig:iou}
\end{figure}

\paragraph{Segmentation Performance}
Following previous works~\citep{Zhou2018TowardsAL,Shi2019DensePC}, we use the mean Intersection-over-Union (mIoU) metric to evaluate the segmentation performance.
Recall that the ground-truth segments are marked by start (and end) times, whereas the predicted segments are expressed according to the position of the corresponding ASR token.
For the \partitiontask task, we compute the token index-based IoU: each ground-truth segment is defined by the start and end ASR token index, and will be compared against the predicted index.
For the \vanilla task, we compute the timestamp-based IoU: predicted indices are mapped into the corresponding ASR token timestamps and compared against the segment's ground-truth start and end timestamps.
Fig.~\ref{fig:iou} provides an example of the two types of IoU used in this study.

An IoU score can be computed for each (ground-truth, predicted) segment pair.
The mIoU measure provides a summary score for segmentation performance over the entire video. For each ground-truth segment, we take its maximal IoU to predicted segments as the IoU score for this ground-truth segment, and mIoU is the average of this value across all ground-truth segments.
The individual mIoU for each video is then averaged across the test data and reported as the overall mIoU.

For diagnostic purposes, we also compute: 1) the percentage of predicted segments which have an IoU score with at least one ground-truth segment above a certain threshold  $t$ (precision@$t$); 2) the percentage of ground-truth segments which have an IoU score with at least one predicted segment, above a certain threshold $t$ (recall@$t$), as well as their geometric mean as F1. Following prior work, we compute these scores for a set of IoU thresholds $t$=\{0.3, 0.5, 0.7, 0.9\}, and report the average over these thresholds.

\paragraph{Captioning Performance}
We compute BLEU-4~\citep{Papineni2002BleuAM}, METEOR~\citep{Banerjee2005METEORAA}, CIDEr~\citep{Vedantam2015CIDErCI}, and ROUGE~\citep{Lin2004ROUGEAP} scores between generated captions and the ground truth when the predicted and ground-truth segment ``match'' (i.e., with IoU score above a given threshold $t$); if a ground truth segment does not have a matching prediction, it contributes a zero to the average score for the corresponding threshold. Again, we compute this for a set of IoU thresholds of \{0.3, 0.5, 0.7, 0.9\}, and report the average over these thresholds.

\subsection{Implementation Details}
Models were trained on 4x4 TPUs, and we used about 180k GPU hours for around 1380 training runs, including pretraining the WikiHow checkpoint, pilot studies with toy examples, debugging, and hyperparameter tuning. The models have approximately 70 million parameters. 
We used the Adafactor \cite{shazeer:18} optimizer and a learning rate schedule of 1000 warmup steps followed by square-root decay. We did a few initial exploratory runs over base learning rates of \{0.001, 0.01, 0.05, 0.1, 0.2, 0.5, 0.8, 1, 2, 5\} to determine that a base learning rate of 1 worked well and used it for all the experiments reported.

For our visual representations, we computed 3D CNN features pre-trained on the Kinetics~\cite{Carreira2017QuoVA, Kay2017TheKH} dataset for frames sampled at 30 fps, resulting in one feature for each 1 second clip.


\begin{table*}[tbp]
\begin{adjustbox}{width=0.98\linewidth,center}
\begin{tabular}{c | r | l | l |  r  | r r r r r}
\cmidrule[\heavyrulewidth]{1-10}
\multirow{2}{*}{\textbf{Dataset}} &  \multirow{2}{*}{\textbf{*}} &  \multirowcell{2}{\textbf{Input}\\\textbf{Formulation}}    & \multirow{2}{*}{\textbf{Checkpoint?}}  &    \multicolumn{1}{c}{\textbf{\segonly}}  &   \multicolumn{5}{|c}{\textbf{\segcap}} \\
\cmidrule{5-10}
    &   &   &    &  mIoU   &  mIoU   &   B@4    &   METEOR  &   CIDEr   &   ROUGE-L \\ 
\cmidrule[\heavyrulewidth]{1-10}

\multirow{9}{*}{Youcook2}
    &   \texttt{0}   & \multicolumn{2}{c|}{Random Segmentation}    & 	20.61	$\pm$	1.04 &   -   &   -   &   -      &   -      &   -        \\
\cmidrule{3-10}
    &   \texttt{1}   &\multirow{4}{*}{\notemporal}
    &   -                & 	12.99	$\pm$	1.55 & 	16.45	$\pm$	8.72 & 	0.17	$\pm$	0.11 & 	0.66	$\pm$	0.04 & 	0.02	$\pm$	0.01 & 	1.99	$\pm$	0.20 \\
    &   \texttt{2}   &    &   T5         & 	24.14	$\pm$	1.07 & 	24.21	$\pm$	1.64 & 	0.88	$\pm$	0.04 & 	1.50	$\pm$	0.12 & 	0.09	$\pm$	0.01 & 	3.34	$\pm$	0.27 \\
    &   \texttt{3}    &    &   WikiHow  & 	22.58	$\pm$	1.09 & 	23.33	$\pm$	0.79 & 	0.67	$\pm$	0.15 & 	1.47	$\pm$	0.05 & 	0.08	$\pm$	0.01 & 	3.51	$\pm$	0.13 \\
    &   \texttt{4}    &    &   WikiHow T5   & 	\textbf{27.77}	$\pm$	0.09 & 	\textbf{30.26}	$\pm$	1.24 & 	\textbf{2.96}	$\pm$	0.28 & 	\textbf{3.49}	$\pm$	0.30 & 	\textbf{0.25}	$\pm$	0.03 & 	\textbf{7.00}	$\pm$	0.42 \\
\cmidrule{3-10}

    &   \texttt{5}    &\multirow{4}{*}{\withtemporal}
    &   -               & 	18.51	$\pm$	1.95 & 	18.71	$\pm$	0.17 & 	0.12	$\pm$	0.07 & 	0.48	$\pm$	0.08 & 	0.02	$\pm$	0.01 & 	1.41	$\pm$	0.22 \\
    &   \texttt{6}    &    &   T5       & 	23.02	$\pm$	1.05 & 	23.96	$\pm$	0.08 & 	1.32	$\pm$	0.08 & 	1.91	$\pm$	0.07 & 	0.11	$\pm$	0.01 & 	4.20	$\pm$	0.13 \\
    &   \texttt{7}    &    &   WikiHow      & 	21.68	$\pm$	1.93 & 	21.88	$\pm$	0.86 & 	0.69	$\pm$	0.19 & 	1.30	$\pm$	0.07 & 	0.07	$\pm$	0.01 & 	3.06	$\pm$	0.13 \\
    &   \texttt{8}    &    &   WikiHow T5    & 	26.51	$\pm$	0.45 & 	28.70	$\pm$	0.92 & 	2.58	$\pm$	0.19 & 	3.23	$\pm$	0.10 & 	0.22	$\pm$	0.01 & 	6.45	$\pm$	0.17 \\

\cmidrule[\heavyrulewidth]{1-10}
\multirow{9}{*}{ViTT}
    &   \texttt{9}   & \multicolumn{2}{c|}{Random Segmentation}    & 	21.90	$\pm$	0.15 & 	   -   &   -   &   -      &   -      &   -        \\
\cmidrule{3-10}
    &   \texttt{10}   &\multirow{4}{*}{\notemporal}
    &   -                & 	33.85	$\pm$	0.70 & 	32.69	$\pm$	0.71 & 	0.11	$\pm$	0.01 & 	3.76	$\pm$	0.35 & 	0.08	$\pm$	0.01 & 	3.86	$\pm$	0.28 \\
    &   \texttt{11}    &    &   T5       & 	37.89	$\pm$	0.10 & 	38.07	$\pm$	0.65 & 	0.57	$\pm$	0.03 & 	5.92	$\pm$	0.37 & 	0.16	$\pm$	0.02 & 	6.59	$\pm$	0.69 \\
    &   \texttt{12}    &    &   WikiHow  & 	38.20	$\pm$	0.27 & 	37.80	$\pm$	0.62 & 	0.40	$\pm$	0.07 & 	5.48	$\pm$	0.18 & 	0.14	$\pm$	0.01 & 	6.02	$\pm$	0.34 \\
    &   \texttt{13}    &    &   WikiHow T5  & 	\textbf{41.87}	$\pm$	0.26 & 	42.40	$\pm$	0.30 & 	\textbf{1.29}	$\pm$	0.07 & 	\textbf{8.10}	$\pm$	0.34 & 	\textbf{0.25}	$\pm$	0.01 & 	\textbf{9.26}	$\pm$	0.39 \\
\cmidrule{3-10}

    &   \texttt{14}    &\multirow{4}{*}{\withtemporal}
    &   -                & 	33.89	$\pm$	0.21 & 	35.37	$\pm$	3.18 & 	0.04	$\pm$	0.03 & 	3.42	$\pm$	0.61 & 	0.07	$\pm$	0.01 & 	3.28	$\pm$	0.83 \\
    &   \texttt{15}    &    &   T5      & 	37.78	$\pm$	0.15 & 	38.50	$\pm$	0.55 & 	0.75	$\pm$	0.10 & 	6.37	$\pm$	0.39 & 	0.18	$\pm$	0.01 & 	7.19	$\pm$	0.48 \\
    &   \texttt{16}    &    &   WikiHow  & 	37.27	$\pm$	0.08 & 	36.97	$\pm$	0.48 & 	0.38	$\pm$	0.06 & 	5.31	$\pm$	0.06 & 	0.13	$\pm$	0.01 & 	5.82	$\pm$	0.23 \\
    &   \texttt{17}    &    &   WikiHow T5  & 	\textbf{41.64}	$\pm$	0.12 & 	\textbf{43.22}	$\pm$	0.72 & 	\textbf{1.22}	$\pm$	0.08 & 	\textbf{8.05}	$\pm$	0.20 & 	\textbf{0.25}	$\pm$	0.01 & 	\textbf{9.18}	$\pm$	0.45 \\

\cmidrule[\heavyrulewidth]{1-10}
\end{tabular}
\end{adjustbox}
\caption{Dense video captioning performance on YouCook2 and ViTT test sets with the \offsetbased formulation. We ran 3 trials for each setting, and report the evaluation results (mean $\pm$ std) with models initialized from random weights, T5 checkpoints, WikiHow checkpoints, and T5 checkpoints further pretrained on WikiHow. Note: Seg stands for the segmentation task, and Cap stands for the captioning task.}
\label{tab:vanilla_dvc_results_mean_std}
\vspace{-1ex}
\end{table*}

\subsection{Experiments in the \partitionsetting}
\paragraph{Experimental setup}
We conduct comparisons of the two different target formulations, \hardencoding and \offsetbased, in the \partitionsetting, using the following experimental setup: (a) max input text length and target length are set to 1024, and max input visual feature length is set to 800; this can truncate longer ASR sequences, but allow us to quickly iterate through different settings with fewer computational resources; 
(b) only one run for each setting.
We report results on the validation set in Table~\ref{tab:youcook2_partition}.

\paragraph{Target formulations}
The best performing model (row \#\texttt{4} in Table~\ref{tab:youcook2_partition}: \offsetbased with T5 checkpoint) outperforms a random partition baseline\footnote{For the random partition baseline, a video is randomly partitioned into $n$ segments, with $n$ sampled uniformly from 1 to 15 (The mean number of segments in the ground-truth is 8).} (row \#\texttt{0} in Table~\ref{tab:youcook2_partition}), 
which indicates our target formulation approach to the segmentation task is capturing some segmentation information effectively. 

When trained from scratch, the \offsetbased formulation achieves higher performance across the board (\#\texttt{3} vs \#\texttt{1}), with a smaller gap for the \segonly, and a more marked lead for the \segcap.
We hypothesize that while treating the segmentation task as a tagging task is more or less feasible on its own, combining segmentation tags and captions is not a suitable formulation for the combined task -- to the point that the \segcap underperforms \segonly in segmentation metrics (mIoU of 19.72 vs. 33.59 in \#\texttt{1}).

The \offsetbased formulation overall benefits from the T5 checkpoint (\#\texttt{3} vs \#\texttt{4} in Table~\ref{tab:youcook2_partition}) across different sub-tasks.  Note that for the \segonly, the target strings (sequences of numbers) are not typically seen in T5 pretraining, but the T5 checkpoint still boosts its performance.
In contrast, the \hardencoding formulation is not able to benefit from the T5 checkpoint in our experiments. One possible explanation is that the target strings in \hardencoding (with large chunks of padding tokens) are just too different from the T5 pretraining targets.  

Given the results obtained in the \partitionsetting, we focus our efforts on using \offsetbased target formulation in the more challenging \vanilla setting in the following section.

\paragraph{Ablation studies} We conducted ablation studies on input modalities, and find models that take text-only inputs stand to benefit more from the pre-trained checkpoints than the models that only take visual inputs.
We also conducted ablation studies on partial parameter initialization, and found that partially loading checkpoints from pre-trained models does not work as well as fully loading checkpoints for both the encoder and the decoder. See Appendix (\ref{sec:appendix:ablation}) for more details.

\subsection{Experiments in the \vanilla setting}
\paragraph{Experimental setup}
Using the \offsetbased target formulation, we conduct a more extensive comparison of the effect of different pretraining strategies, as well as different input formulations on the \vanilla dense video captioning task on both YouCook2 and ViTT.
Maximum sequence lengths are set to ensure no truncation happens in either dataset -- input text: 4096; visual feature: 800 (YouCook2) / 500 (ViTT); target: 512 (YouCook2) / 256 (ViTT).
We ran each experiment with different seeds three times to account for performance variance from random initializations.  
We report the mean and standard deviation (using 3 runs) for each metric in Table \ref{tab:vanilla_dvc_results_mean_std}.
We choose the best checkpoint based on performance on the validation set and report the performance on the test set.

\paragraph{Effects of Pretraining} For both datasets, there are significant performance improvements from utilizing pre-trained checkpoints in terms of both segmentation metrics and captioning metrics.
Interestingly, training from the WikiHow checkpoint (using in-domain task over in-domain data) provides similar performance improvement to T5 alone (see, for instance, \#\texttt{2} vs \#\texttt{3}, or \#\texttt{11} vs \#\texttt{12} in Table \ref{tab:vanilla_dvc_results_mean_std}).
However, starting from the generic-language T5 checkpoint and adding in-domain WikiHow pretraining (WikiHow T5, e.g., \#\texttt{4} and \#\texttt{13}) boosts all metrics by a large and significant margin.

\paragraph{Effects of Joint Modeling}
If we compare the mIoU score achieved by the \segcap to the mIoU score achieved by the \segonly in Table \ref{tab:vanilla_dvc_results_mean_std},
across different settings,
we observe a general trend where the \segcap outperforms the \segonly on this segmentation metric.
This indicates that with the right formulation, the segmentation subtask (predicting event boundary) can indeed benefit from joint learning with a related captioning subtask (summarizing event content).

\begin{table}[tbp]
\begin{adjustbox}{width=\linewidth,center}
\begin{tabular}{l  r r r r r}
\cmidrule[\heavyrulewidth]{1-6}
\textbf{Model}   & \textbf{mIoU}  & \textbf{Prec.}  & \textbf{Rec.}      &   \textbf{B@4}    &   \textbf{M}   \\ 
\cmidrule[\heavyrulewidth]{1-6}
vsLSTM~\citep{Zhang2016VideoSW}                 & 32.2  & 24.1  & 22.1  & -  & -  \\
SCNN-prop~\citep{Shou2016TemporalAL}            & 26.7  & 23.2  & 28.2  & -  & -  \\
ProcNet~\citep{Zhou2018TowardsAL}               & 37.0  & 30.4  & 37.1  & -  & -  \\
Bi-LSTM + TempoAttn~\citep{Zhou2018EndtoEndDV} & -  & -  & -  & 0.08  & 4.62  \\
End2end Transformer~\citep{Zhou2018EndtoEndDV} & -  & -  & -  & 0.30  & 6.58  \\
Context-aware Fusion~\citep{Shi2019DensePC}& 41.4  & -  & -  & 2.61  & 17.43  \\
\cmidrule{1-6}
End2end Sequence Generation (Ours)    &   30.3 & 	24.5 & 	24.2 & 	2.96 & 	3.49 \\
\cmidrule[\heavyrulewidth]{1-6}
\end{tabular}
\end{adjustbox}
\caption{Dense video captioning performance on YouCook2 in the context of prior work. Following prior work, the segmentation performance is measured by the mIoU, the precision (Prec.) and recall (Rec.) at IoU threshold $t$=0.5.
Captioning performance is measured by the average BLEU-4 (B@4) and METEOR (M) at IoU thresholds $t \in$\{0.3, 0.5, 0.7, 0.9\}.
}
\label{tab:prior_work_youcook2}
\end{table}

\paragraph{Input formulations} 
Results using \SimpleConcatShort compared to their counterparts using \TemporalEmbShort in Table \ref{tab:vanilla_dvc_results_mean_std}, are mixed.
While \TemporalEmbShort seems to bring non-trivial improvement to models trained from scratch, the training from scratch settings also has the largest variance in our experiments\footnote{To the extent that the \segcap performance in \#\texttt{1} can be considered an outlier: its mIoU scores for the three runs are (11, 11, 26), which resulted in a large std value not seen anywhere else in the table.  We looked into these three runs in more details, and our best guess was that one of them incidentally got an advantageous random initialization.}.
That said, the \segcap did achieve its best mIoU score on ViTT using \TemporalEmbShort.
More work is needed to fully understand the potential of \TemporalEmbShort.

\paragraph{Comparison against prior work for YouCook2\footnote{ViTT is a relatively newer dataset and past work has only reported performance of the segment-level captioning subtask using ground-truth segments; we are not aware of existing work reporting end-to-end dense video captioning performance.}}
Table~\ref{tab:prior_work_youcook2} provides a summary of dense video captioning performance on YouCook2 reported in prior work.
Some of the prior work~\citep{Zhang2016VideoSW,Shou2016TemporalAL,Zhou2018TowardsAL} focused only on the segmentation subtask, while some~\citep{Zhou2018EndtoEndDV,Shi2019DensePC} approached the end-to-end task as a two-stage task and solved the two subtasks separately.
In this context, we find the results from our simple end-to-end sequence generation based approach quite encouraging, and hope this inspires future studies to fully realize the potential of this alternative approach.

\paragraph{Qualitative Analysis} 
We provide a few example model outputs from our \segcap.
More examples can be found in the Appendix~\ref{sec:appendix:showcases}.

Figure~\ref{fig:seg_eg} presents example segmentation results.  
As reflected in Figure~\ref{fig:seg_eg}(a), a segmentation prediction that is largely correct for a few segments, but is missing out on some ground-truth segments and contains over-segmentation of others can result in a relatively low mIoU score. 
For examples with relatively high mIoU scores, see Figure~\ref{fig:seg_eg}(b) (taken from the more challenging YouCook2 setting, with gaps between ground-truth segments), and Figure~\ref{fig:seg_eg}(c) (taken from ViTT, with no gaps between segments).

Next, we observe that caption quality is good when the segment boundary prediction is highly accurate:
if we restrict to segments with IoU $\ge 90\%$ between the prediction and the target, the average ROUGE-L score for corresponding captions is 30.18 for YouCook2 and 44.33 for ViTT.
Table~\ref{tab:caption_example} presents qualitative examples.

\begin{figure*}[tbp]
    \centering
    \resizebox{0.65\linewidth}{!}{%
    \includegraphics{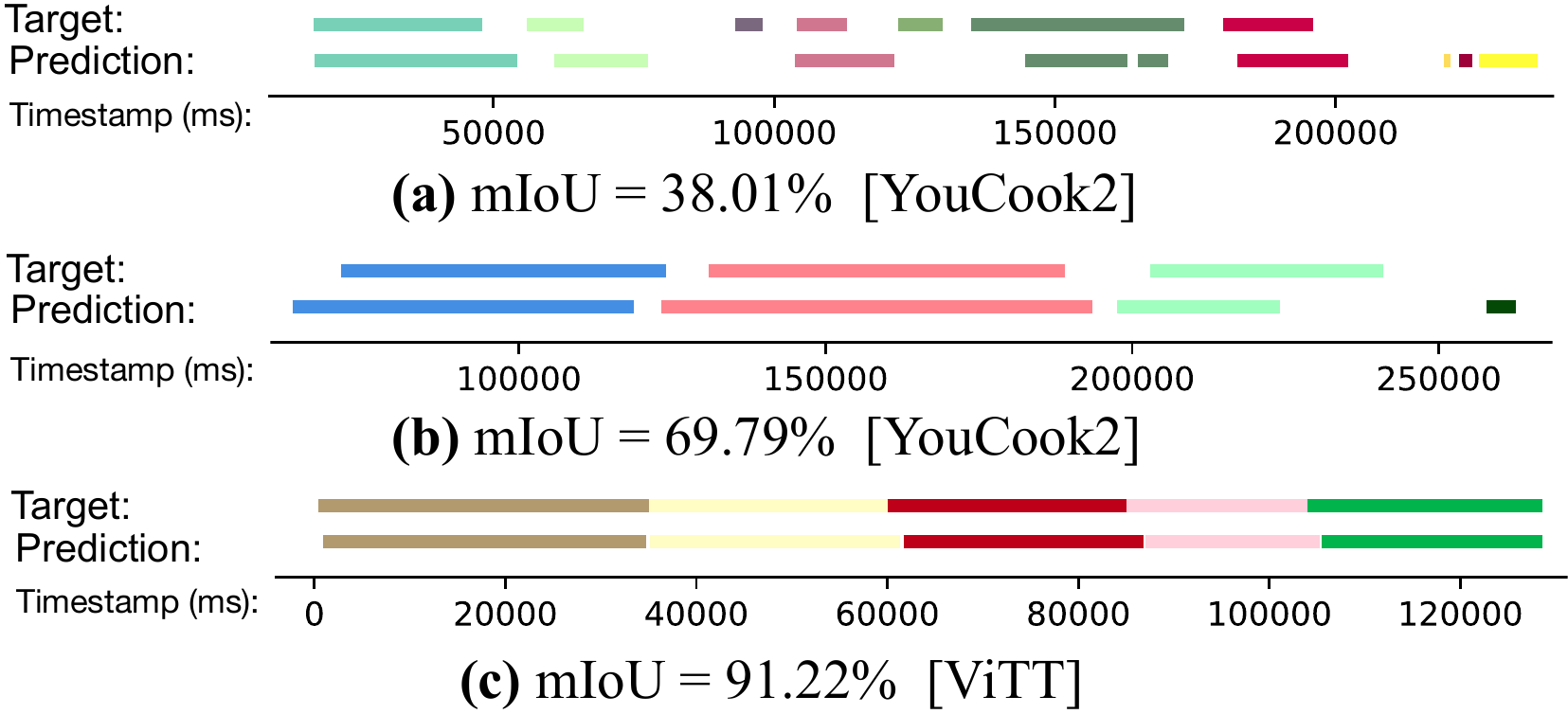}
    }
    \caption{Example segmentation predictions corresponding to different mIoU scores.
}
    \label{fig:seg_eg}
\end{figure*}

\begin{table*}[tbp]
\begin{adjustbox}{width=0.7\linewidth,center}
\begin{tabular}{l  r r l}
\cmidrule[\heavyrulewidth]{1-4}
   & \textbf{IoU}  & \textbf{Segment Border (ms)}  & \textbf{Caption} \\ 
\cmidrule[\heavyrulewidth]{1-4}
Tgt.      &  \multirow{2}{*}{90.0\%}     &   [58000.0, 77000.0]	& whisk eggs and season with salt \\
Pred. 	&   &   [57309.0, 78429.5]	& whisk the eggs in the deep plate \\
\cmidrule{1-4}
Tgt.    &   \multirow{2}{*}{99.3\%}  &    	[28000.0, 45000.0]  &	chop up the garlic in the food processer \\
Pred.   &   & 	[28005.0, 44894.0]  &	chop garlic and place in the food processor \\
\cmidrule[\heavyrulewidth]{1-4}
Tgt.  &  \multirow{2}{*}{94.7\%} &	[64199.0, 98080.0]	&   Preparining remaining ingredients \\
Pred. & &	[65710.0, 98380.0]	&   Chopping the remaining ingredients \\
\cmidrule{1-4}
Tgt.    &   \multirow{2}{*}{97.0\%}    &	[65100.0, 124729.0]	&   Blow-drying the roots \\
Pred.   &   & 	[63239.5, 124714.5]	    &   Blow-drying hair \\
\cmidrule[\heavyrulewidth]{1-4}
\end{tabular}
\end{adjustbox}
\caption{Example caption predictions where the IoU $\ge 90\%$ between the target (Tgt.) and the predicted (Pred.) segments. The first two examples are from YouCook2, the last two examples are from ViTT.
}
\label{tab:caption_example}
\vspace{-2ex}
\end{table*}

\section{Conclusion}

In this paper, we describe different task formulations for solving the dense video captioning task using an end-to-end sequence generation approach, which allows us to leverage pre-trained text-only encoder-decoder models. 
We conduct experiments on YouCook2 and ViTT in several pretraining settings.
Experimental results show that general (T5) and in-domain (WikiHow) text-only pre-trained models both improve video partitioning and segmentation performance, and the gains are cumulative.
Also, the segmentation subtask benefits from joint modeling with the captioning subtask.
We hope our work can inspire future studies on leveraging pre-trained models, large-scale text corpora and language generation formulations to solve multimodal tasks such as dense video captioning.


\section*{Ethical Statement}
Our experiments are conducted only on videos with available English ASR annotations, as we inherit this limitation from the available data for this task.
We use existing datasets based on public YouTube videos.
As a consequence, any videos that are no longer publicly available on YouTube (e.g., removed by the user) at the time of the study needed to be excluded from our experimental setup.


\bibliography{anthology,custom}

\begin{thebibliography}{48}
\expandafter\ifx\csname natexlab\endcsname\relax\def\natexlab#1{#1}\fi

\bibitem[{Abu-El-Haija et~al.(2016)Abu-El-Haija, Kothari, Lee, Natsev,
  Toderici, Varadarajan, and Vijayanarasimhan}]{AbuElHaija2016YouTube8MAL}
Sami Abu-El-Haija, Nisarg Kothari, Joonseok Lee, Apostol Natsev, George
  Toderici, Balakrishnan Varadarajan, and Sudheendra Vijayanarasimhan. 2016.
\newblock Youtube-8m: A large-scale video classification benchmark.
\newblock \emph{ArXiv}, abs/1609.08675.

\bibitem[{Agrawal et~al.(2015)Agrawal, Lu, Antol, Mitchell, Zitnick, Parikh,
  and Batra}]{Agrawal2015VQAVQ}
Aishwarya Agrawal, Jiasen Lu, Stanislaw Antol, Margaret Mitchell, C.~Lawrence
  Zitnick, Devi Parikh, and Dhruv Batra. 2015.
\newblock Vqa: Visual question answering.
\newblock \emph{International Journal of Computer Vision}, 123:4--31.

\bibitem[{Baheti(2019)}]{baheti_2019}
Prashant Baheti. 2019.
\newblock \href
  {https://www.blog.google/products/search/key-moments-video-search/} {Search
  helps you find key moments in videos}.

\bibitem[{Banerjee and Lavie(2005)}]{Banerjee2005METEORAA}
Satanjeev Banerjee and Alon Lavie. 2005.
\newblock Meteor: An automatic metric for mt evaluation with improved
  correlation with human judgments.
\newblock In \emph{IEEvaluation@ACL}.

\bibitem[{Carreira and Zisserman(2017)}]{Carreira2017QuoVA}
Jo{\~a}o Carreira and Andrew Zisserman. 2017.
\newblock Quo vadis, action recognition? a new model and the kinetics dataset.
\newblock \emph{2017 IEEE Conference on Computer Vision and Pattern Recognition
  (CVPR)}, pages 4724--4733.

\bibitem[{Chen et~al.(2022)Chen, Saxena, Li, Fleet, and
  Hinton}]{chen2022pixseq}
Ting Chen, Saurabh Saxena, Lala Li, David~J. Fleet, and Geoffrey Hinton. 2022.
\newblock Pix2seq: A language modeling framework for object detection.
\newblock In \emph{International Conference on Learning Representations}.

\bibitem[{Chen et~al.(2015)Chen, Fang, Lin, Vedantam, Gupta, Doll{\'a}r, and
  Zitnick}]{cococap}
Xinlei Chen, Hao Fang, Tsung-Yi Lin, Ramakrishna Vedantam, Saurabh Gupta, Piotr
  Doll{\'a}r, and C.~Lawrence Zitnick. 2015.
\newblock {Microsoft COCO Captions}: Data collection and evaluation server.
\newblock \emph{arXiv preprint arXiv:1504.00325}.

\bibitem[{Chen et~al.(2020)Chen, Li, Yu, Kholy, Ahmed, Gan, Cheng, and
  Liu}]{Chen2020UNITERUI}
Yen-Chun Chen, Linjie Li, Licheng Yu, Ahmed~El Kholy, Faisal Ahmed, Zhe Gan,
  Yu~Cheng, and Jingjing Liu. 2020.
\newblock Uniter: Universal image-text representation learning.
\newblock In \emph{ECCV}.

\bibitem[{Deng et~al.(2021)Deng, Chen, Chen, He, and Wu}]{Deng2021SketchGA}
Chaorui Deng, Shizhe Chen, Da~Chen, Yuan He, and Qi~Wu. 2021.
\newblock Sketch, ground, and refine: Top-down dense video captioning.
\newblock \emph{2021 IEEE/CVF Conference on Computer Vision and Pattern
  Recognition (CVPR)}, pages 234--243.

\bibitem[{Devlin et~al.(2019)Devlin, Chang, Lee, and
  Toutanova}]{Devlin2019BERTPO}
Jacob Devlin, Ming-Wei Chang, Kenton Lee, and Kristina Toutanova. 2019.
\newblock Bert: Pre-training of deep bidirectional transformers for language
  understanding.
\newblock In \emph{NAACL}.

\bibitem[{Gan et~al.(2020)Gan, Chen, Li, Zhu, Cheng, and
  Liu}]{Gan2020LargeScaleAT}
Zhe Gan, Yen-Chun Chen, Linjie Li, Chen Zhu, Yu~Cheng, and Jingjing Liu. 2020.
\newblock Large-scale adversarial training for vision-and-language
  representation learning.
\newblock \emph{ArXiv}, abs/2006.06195.

\bibitem[{Huang et~al.(2020{\natexlab{a}})Huang, Pang, Zhu, Rivera, and
  Soricut}]{Huang2020MultimodalPF}
Gabriel Huang, Bo~Pang, Zhenhai Zhu, Clara Rivera, and Radu Soricut.
  2020{\natexlab{a}}.
\newblock Multimodal pretraining for dense video captioning.
\newblock In \emph{AACL}.

\bibitem[{Huang et~al.(2020{\natexlab{b}})Huang, Zeng, Liu, Fu, and
  Fu}]{Huang2020PixelBERTAI}
Zhicheng Huang, Zhaoyang Zeng, Bei Liu, Dongmei Fu, and Jianlong Fu.
  2020{\natexlab{b}}.
\newblock Pixel-bert: Aligning image pixels with text by deep multi-modal
  transformers.
\newblock \emph{ArXiv}, abs/2004.00849.

\bibitem[{Iashin and Rahtu(2020)}]{Iashin2020MultimodalDV}
Vladimir Iashin and Esa Rahtu. 2020.
\newblock Multi-modal dense video captioning.
\newblock \emph{2020 IEEE/CVF Conference on Computer Vision and Pattern
  Recognition Workshops (CVPRW)}, pages 4117--4126.

\bibitem[{Kay et~al.(2017)Kay, Carreira, Simonyan, Zhang, Hillier,
  Vijayanarasimhan, Viola, Green, Back, Natsev, Suleyman, and
  Zisserman}]{Kay2017TheKH}
Will Kay, Jo{\~a}o Carreira, Karen Simonyan, Brian Zhang, Chloe Hillier,
  Sudheendra Vijayanarasimhan, Fabio Viola, Tim Green, Trevor Back, Apostol
  Natsev, Mustafa Suleyman, and Andrew Zisserman. 2017.
\newblock The kinetics human action video dataset.
\newblock \emph{ArXiv}, abs/1705.06950.

\bibitem[{Kim et~al.(2021)Kim, Son, and Kim}]{Kim2021ViLTVT}
Wonjae Kim, Bokyung Son, and Ildoo Kim. 2021.
\newblock Vilt: Vision-and-language transformer without convolution or region
  supervision.
\newblock In \emph{ICML}.

\bibitem[{Koupaee and Wang(2018)}]{Koupaee2018WikiHowAL}
Mahnaz Koupaee and William~Yang Wang. 2018.
\newblock Wikihow: A large scale text summarization dataset.
\newblock \emph{ArXiv}, abs/1810.09305.

\bibitem[{Krishna et~al.(2017)Krishna, Hata, Ren, Fei-Fei, and
  Niebles}]{Krishna2017DenseCaptioningEI}
Ranjay Krishna, Kenji Hata, Frederic Ren, Li~Fei-Fei, and Juan~Carlos Niebles.
  2017.
\newblock Dense-captioning events in videos.
\newblock \emph{2017 IEEE International Conference on Computer Vision (ICCV)},
  pages 706--715.

\bibitem[{Li et~al.(2020{\natexlab{a}})Li, Duan, Fang, Jiang, and
  Zhou}]{Li2020UnicoderVLAU}
Gen Li, Nan Duan, Yuejian Fang, Daxin Jiang, and Ming Zhou. 2020{\natexlab{a}}.
\newblock Unicoder-vl: A universal encoder for vision and language by
  cross-modal pre-training.
\newblock In \emph{AAAI}.

\bibitem[{Li et~al.(2019)Li, Yatskar, Yin, Hsieh, and
  Chang}]{Li2019VisualBERTAS}
Liunian~Harold Li, Mark Yatskar, Da~Yin, Cho-Jui Hsieh, and Kai-Wei Chang.
  2019.
\newblock Visualbert: A simple and performant baseline for vision and language.
\newblock \emph{ArXiv}, abs/1908.03557.

\bibitem[{Li et~al.(2020{\natexlab{b}})Li, Yin, Li, Hu, Zhang, Zhang, Wang, Hu,
  Dong, Wei, Choi, and Gao}]{Li2020OscarOA}
Xiujun Li, Xi~Yin, Chunyuan Li, Xiaowei Hu, Pengchuan Zhang, Lei Zhang, Lijuan
  Wang, Houdong Hu, Li~Dong, Furu Wei, Yejin Choi, and Jianfeng Gao.
  2020{\natexlab{b}}.
\newblock Oscar: Object-semantics aligned pre-training for vision-language
  tasks.
\newblock In \emph{ECCV}.

\bibitem[{Li et~al.(2018)Li, Yao, Pan, Chao, and Mei}]{Li2018JointlyLA}
Yehao Li, Ting Yao, Yingwei Pan, Hongyang Chao, and Tao Mei. 2018.
\newblock Jointly localizing and describing events for dense video captioning.
\newblock \emph{2018 IEEE/CVF Conference on Computer Vision and Pattern
  Recognition}, pages 7492--7500.

\bibitem[{Lin(2004)}]{Lin2004ROUGEAP}
Chin-Yew Lin. 2004.
\newblock Rouge: A package for automatic evaluation of summaries.
\newblock In \emph{ACL 2004}.

\bibitem[{Lin et~al.(2020)Lin, Yang, Zhang, Liu, Zhou, and
  Yang}]{Lin2020InterBERTVI}
Junyang Lin, An~Yang, Yichang Zhang, Jianbin Liu, Jingren Zhou, and Hongxia
  Yang. 2020.
\newblock Interbert: Vision-and-language interaction for multi-modal
  pretraining.
\newblock \emph{ArXiv}, abs/2003.13198.

\bibitem[{Lu et~al.(2019)Lu, Batra, Parikh, and Lee}]{Lu2019ViLBERTPT}
Jiasen Lu, Dhruv Batra, Devi Parikh, and Stefan Lee. 2019.
\newblock Vilbert: Pretraining task-agnostic visiolinguistic representations
  for vision-and-language tasks.
\newblock In \emph{NeurIPS}.

\bibitem[{Luo et~al.(2020)Luo, Ji, Shi, Huang, Duan, Li, Chen, and
  Zhou}]{Luo2020UniViLMAU}
Huaishao Luo, Lei Ji, Botian Shi, Haoyang Huang, Nan Duan, Tianrui Li, Xilin
  Chen, and Ming Zhou. 2020.
\newblock Univilm: A unified video and language pre-training model for
  multimodal understanding and generation.
\newblock \emph{ArXiv}, abs/2002.06353.

\bibitem[{Miech et~al.(2019)Miech, Zhukov, Alayrac, Tapaswi, Laptev, and
  Sivic}]{miech2019howto100m}
Antoine Miech, Dimitri Zhukov, Jean-Baptiste Alayrac, Makarand Tapaswi, Ivan
  Laptev, and Josef Sivic. 2019.
\newblock Howto100m: Learning a text-video embedding by watching hundred
  million narrated video clips.
\newblock \emph{ArXiv}, abs/1906.03327.

\bibitem[{Mun et~al.(2019)Mun, Yang, Ren, Xu, and Han}]{Mun2019StreamlinedDV}
Jonghwan Mun, L.~Yang, Zhou Ren, N.~Xu, and Bohyung Han. 2019.
\newblock Streamlined dense video captioning.
\newblock \emph{2019 IEEE/CVF Conference on Computer Vision and Pattern
  Recognition (CVPR)}, pages 6581--6590.

\bibitem[{O’Neil-Hart(2017)}]{oneil-hart_2017}
Celie O’Neil-Hart. 2017.
\newblock \href
  {https://www.thinkwithgoogle.com/marketing-strategies/video/self-directed-learning-youtube/}
  {Self-directed learning from youtube - think with google}.

\bibitem[{Papineni et~al.(2002)Papineni, Roukos, Ward, and
  Zhu}]{Papineni2002BleuAM}
Kishore Papineni, Salim Roukos, Todd Ward, and Wei-Jing Zhu. 2002.
\newblock Bleu: a method for automatic evaluation of machine translation.
\newblock In \emph{ACL}.

\bibitem[{Qi et~al.(2020)Qi, Su, Song, Cui, Bharti, and
  Sacheti}]{Qi2020ImageBERTCP}
Di~Qi, Lin Su, Jianwei Song, Edward Cui, Taroon Bharti, and Arun Sacheti. 2020.
\newblock Imagebert: Cross-modal pre-training with large-scale weak-supervised
  image-text data.
\newblock \emph{ArXiv}, abs/2001.07966.

\bibitem[{Raffel et~al.(2020)Raffel, Shazeer, Roberts, Lee, Narang, Matena,
  Zhou, Li, and Liu}]{t5}
Colin Raffel, Noam Shazeer, Adam Roberts, Katherine Lee, Sharan Narang, Michael
  Matena, Yanqi Zhou, Wei Li, and Peter~J. Liu. 2020.
\newblock Exploring the limits of transfer learning with a unified text-to-text
  transformer.
\newblock \emph{JMLR}.

\bibitem[{Ramshaw and Marcus(1995)}]{ramshaw-marcus-1995-text}
Lance Ramshaw and Mitch Marcus. 1995.
\newblock Text chunking using transformation-based learning.
\newblock In \emph{Third Workshop on Very Large Corpora}.

\bibitem[{Shazeer and Stern(2018)}]{shazeer:18}
Noam Shazeer and Mitchell Stern. 2018.
\newblock Adafactor: Adaptive learning rates with sublinear memory cost.
\newblock In \emph{Proceedings of the 35th International Conference on Machine
  Learning, {ICML} 2018, Stockholmsm{\"{a}}ssan, Stockholm, Sweden, July 10-15,
  2018}, volume~80 of \emph{Proceedings of Machine Learning Research}, pages
  4603--4611. {PMLR}.

\bibitem[{Shi et~al.(2019)Shi, Ji, Liang, Duan, Chen, Niu, and
  Zhou}]{Shi2019DensePC}
Botian Shi, Lei Ji, Yaobo Liang, Nan Duan, Peng Chen, Zhendong Niu, and
  M.~Zhou. 2019.
\newblock Dense procedure captioning in narrated instructional videos.
\newblock In \emph{ACL}.

\bibitem[{Shou et~al.(2016)Shou, Wang, and Chang}]{Shou2016TemporalAL}
Zheng Shou, Dongang Wang, and Shih-Fu Chang. 2016.
\newblock Temporal action localization in untrimmed videos via multi-stage
  cnns.
\newblock \emph{2016 IEEE Conference on Computer Vision and Pattern Recognition
  (CVPR)}, pages 1049--1058.

\bibitem[{Sun et~al.(2019)Sun, Myers, Vondrick, Murphy, and
  Schmid}]{Sun2019VideoBERTAJ}
Chen Sun, Austin Myers, Carl Vondrick, Kevin~P. Murphy, and Cordelia Schmid.
  2019.
\newblock Videobert: A joint model for video and language representation
  learning.
\newblock \emph{2019 IEEE/CVF International Conference on Computer Vision
  (ICCV)}, pages 7463--7472.

\bibitem[{Tan and Bansal(2019)}]{Tan2019LXMERTLC}
Hao~Hao Tan and Mohit Bansal. 2019.
\newblock Lxmert: Learning cross-modality encoder representations from
  transformers.
\newblock In \emph{EMNLP}.

\bibitem[{Vaswani et~al.(2017)Vaswani, Shazeer, Parmar, Uszkoreit, Jones,
  Gomez, Kaiser, and Polosukhin}]{Vaswani2017AttentionIA}
Ashish Vaswani, Noam~M. Shazeer, Niki Parmar, Jakob Uszkoreit, Llion Jones,
  Aidan~N. Gomez, Lukasz Kaiser, and Illia Polosukhin. 2017.
\newblock Attention is all you need.
\newblock \emph{ArXiv}, abs/1706.03762.

\bibitem[{Vedantam et~al.(2015)Vedantam, Zitnick, and
  Parikh}]{Vedantam2015CIDErCI}
Ramakrishna Vedantam, C.~Lawrence Zitnick, and Devi Parikh. 2015.
\newblock Cider: Consensus-based image description evaluation.
\newblock \emph{2015 IEEE Conference on Computer Vision and Pattern Recognition
  (CVPR)}, pages 4566--4575.

\bibitem[{Wang et~al.(2018)Wang, Jiang, Ma, Liu, and
  Xu}]{Wang2018BidirectionalAF}
Jingwen Wang, Wenhao Jiang, Lin Ma, W.~Liu, and Yong Xu. 2018.
\newblock Bidirectional attentive fusion with context gating for dense video
  captioning.
\newblock \emph{2018 IEEE/CVF Conference on Computer Vision and Pattern
  Recognition}, pages 7190--7198.

\bibitem[{Wang et~al.(2021)Wang, Zhang, Lu, Zheng, Cheng, and
  Luo}]{Wang2021EndtoEndDV}
Teng Wang, Ruimao Zhang, Zhichao Lu, Feng Zheng, Ran Cheng, and Ping Luo. 2021.
\newblock End-to-end dense video captioning with parallel decoding.
\newblock \emph{ArXiv}, abs/2108.07781.

\bibitem[{Xie et~al.(2019)Xie, Lai, Doran, and Kadav}]{Xie2019VisualEA}
Ning Xie, Farley Lai, Derek Doran, and Asim Kadav. 2019.
\newblock Visual entailment: A novel task for fine-grained image understanding.
\newblock \emph{ArXiv}, abs/1901.06706.

\bibitem[{Zellers et~al.(2019)Zellers, Bisk, Farhadi, and
  Choi}]{Zellers2019FromRT}
Rowan Zellers, Yonatan Bisk, Ali Farhadi, and Yejin Choi. 2019.
\newblock From recognition to cognition: Visual commonsense reasoning.
\newblock \emph{2019 IEEE/CVF Conference on Computer Vision and Pattern
  Recognition (CVPR)}, pages 6713--6724.

\bibitem[{Zhang et~al.(2016)Zhang, Chao, Sha, and Grauman}]{Zhang2016VideoSW}
Ke~Zhang, Wei-Lun Chao, Fei Sha, and Kristen Grauman. 2016.
\newblock Video summarization with long short-term memory.
\newblock In \emph{ECCV}.

\bibitem[{Zhou et~al.(2018{\natexlab{a}})Zhou, Xu, and Corso}]{ZhXuCoAAAI18}
Luowei Zhou, Chenliang Xu, and Jason~J Corso. 2018{\natexlab{a}}.
\newblock Towards automatic learning of procedures from web instructional
  videos.
\newblock In \emph{AAAI Conference on Artificial Intelligence}, pages
  7590--7598.

\bibitem[{Zhou et~al.(2018{\natexlab{b}})Zhou, Xu, and
  Corso}]{Zhou2018TowardsAL}
Luowei Zhou, Chenliang Xu, and Jason~J. Corso. 2018{\natexlab{b}}.
\newblock Towards automatic learning of procedures from web instructional
  videos.
\newblock In \emph{AAAI}.

\bibitem[{Zhou et~al.(2018{\natexlab{c}})Zhou, Zhou, Corso, Socher, and
  Xiong}]{Zhou2018EndtoEndDV}
Luowei Zhou, Yingbo Zhou, Jason~J. Corso, Richard Socher, and Caiming Xiong.
  2018{\natexlab{c}}.
\newblock End-to-end dense video captioning with masked transformer.
\newblock \emph{2018 IEEE/CVF Conference on Computer Vision and Pattern
  Recognition}, pages 8739--8748.

\end{thebibliography}
\bibliographystyle{acl_natbib}

\clearpage

\appendix

\section{Appendix}
\label{sec:appendix}

\subsection{Timestamp markers (\timeanchorShort)}
\label{sec:appendix:markers}

Here we describe an alternative way to encode temporal alignment between textual and visual input.
Since the frames are extracted at a fixed rate,  we can explicitly add time markers to the text input to ``mark'' out tokens spoken at the corresponding time points.
The video features are extracted with a frame rate of 1 frame per second in our work.  
We insert a time marker for each frame after the last ASR token spoke before the corresponding timestamp. A time marker consists of a special anchor token, followed by the timestamp token (an integer corresponding to the timestamp in seconds).

Performance using this input formuation can be found in the \timeanchorShort rows in Table \ref{tab:vanilla_dvc_results_mean_std_full}.
For models trained from scratch, including the timestamp markers can positively impact model performance, indicating that these markers do indeed provide helpful information. However, adding these markers only hurt the performance of any model trained from an existing checkpoint. We hypothesize that this is because the text sequence with frequent markers is too different from the pre-trained datasets, leaving the pre-trained checkpoints less effective for models using this input formulation.

\subsection{Ablation studies}
\label{sec:appendix:ablation}

\paragraph{Ablation on Input Data}
Table~\ref{tab:ablation_input_source} shows comparisons of different input sources on YouCook2 dense video captioning task. For all three settings (text-only, video-only, text+video), pre-training on WikiHow has the best performance on both subtasks, and using the T5 checkpoint has better performance than training from scratch. With the pre-trained WikiHow checkpoint, the ``Text-only'' setting has comparable performance as the ``Text+Video'' setting that takes both the ASR transcript and the video features as input. Using the video features alone results in worse performance, indicating the high value of text transcripts to the captioning task.


\paragraph{Ablation on T5 Checkpoint}

Table~\ref{tab:ablation_t5_ckpt} compares performances when using different pre-trained checkpoints on YouCook2 \partitiontask. Using either the T5 or the WikiHow T5 checkpoints outperforms the model initialized from random weights, which verifies the effectiveness of pre-training. Since the targets in the end task are markedly different from, say, T5 pretraining targets, we also experimented with loading partial checkpoints (e.g., only encoder weights). Interestingly, using the full checkpoint has better performance than loading only encoder or only decoder weights.

\begin{table*}[tbp]
\begin{adjustbox}{width=0.8\linewidth,center}
\begin{tabular}{l  l | r r r | r r r r r}
\cmidrule[\heavyrulewidth]{1-10}
\multirow{2}{*}{\textbf{Input}}    & \multirow{2}{*}{\textbf{Checkpoint}}  &    \multicolumn{3}{c}{\textbf{Segmentation}}  &   \multicolumn{5}{|c}{\textbf{Segmentation + Captioning}} \\
\cmidrule{3-10}
   &    &  Precision  & Recall    &   F1  &     F1  &   B@4    &   METEOR  &   CIDEr   &   ROUGE-L \\ 
\cmidrule[\heavyrulewidth]{1-10}

\multirow{3}{*}{Text-only}
    &   -               & 31.86	 & 	30.66	 & 	31.25	 & 	30.39	 & 	0.55	 & 	1.88	 & 	0.07	 & 	5.23\\ 
    &   T5              & 36.22	 & 	37.06	 & 	36.64	 & 	37.89	 & 	3.36	 & 	4.76	 & 	0.28	 & 	10.61\\ 
    &   WikiHow T5      & \textbf{71.13}	 & 	\textbf{63.77}	 & 	\textbf{67.25}	 & 	\textbf{58.71}	 & 	9.57	 & 	\textbf{11.99}	 & 	0.85	 & 	23.21\\ 
\cmidrule{1-10}
	
\multirow{3}{*}{Video-only}
    &   -               & 28.02	 & 	19.48	 & 	22.98	 & 	27.5	 & 	0.52	 & 	1.89	 & 	0.07	 & 	4.82\\ 
    &   T5              & 27.43	 & 	27.25	 & 	27.34	 & 	27.86	 & 	0.40	 & 	1.65	 & 	0.05	 & 	4.11\\ 
    &   WikiHow T5      	 & 25.45	 & 	24.93	 & 	25.19	 & 	23.19	 & 	0.42	 & 	1.48	 & 	0.05	 & 	3.84\\ 
\cmidrule{1-10}

\multirow{3}{*}{Text + Video}
    &   -               & 32.53	 & 	30.90	 & 	31.69	 & 	29.09	 & 	0.34	 & 	1.68	 & 	0.06	 & 	4.78\\ 
    &   T5              & 36.96	 & 	37.99	 & 	37.47	 & 	32.58	 & 	2.99	 & 	4.22	 & 	0.26	 & 	9.20\\ 
    &   WikiHow T5      	 & 71.07	 & 	62.76	 & 	66.66	 & 	57.84	 & 	\textbf{9.87}	 & 	11.96	 & 	\textbf{0.86}	 & 	\textbf{23.25}\\ 
\cmidrule[\heavyrulewidth]{1-10}
\end{tabular}
\end{adjustbox}
\caption{Ablation on input modalities.  Performance using \offsetbased target formulation on YouCook2 dense video captioning task with IoU threshold=50\%. Results are reported on three ablated input settings: ``Text-only'' feeds in the ASR tokens,  ``Video-only'' reveals the video features, while ``Text+Video'' provides both the ASR and the video features as input.}
\label{tab:ablation_input_source}
\end{table*}

\begin{table*}[tbp]
\begin{adjustbox}{width=0.6\linewidth,center}
\begin{tabular}{l | r r r r r }
\cmidrule[\heavyrulewidth]{1-6}
\textbf{Checkpoint}  &   \textbf{F1}  &   \textbf{B@4}    &   \textbf{METEOR}  &   \textbf{CIDEr}   &   \textbf{ROUGE-L} \\ 
\cmidrule[\heavyrulewidth]{1-6}

-                       &  30.39  & 0.55  & 1.88  & 0.07  & 5.23 \\
\cmidrule{1-6}

T5 (full)               &  37.89 & 3.36    & 4.76    & 0.28    & 10.61 \\
T5 (enc-only)           &  31.00    & 0.28    & 1.93    & 0.07    & 4.88 \\
T5 (dec-only)           &  32.37    & 1.39    & 3.02    & 0.14    & 7.73    \\
\cmidrule{1-6}

WikiHow T5 (full)       &  58.71    & \textbf{9.57}    & \textbf{11.99}   & \textbf{0.85}    & \textbf{23.21} \\
WikiHow T5 (enc-only)   &  \textbf{59.30}    & 8.44    & 11.72    & 0.80    & 22.89 \\
WikiHow T5 (dec-only)   &  36.88    & 0.99    & 3.19    & 0.15    & 8.00 \\

\cmidrule[\heavyrulewidth]{1-6}
\end{tabular}
\end{adjustbox}
\caption{Ablation on pretrained checkpoints.  Performance using \offsetbased target formulation on YouCook2 \partitiontask with IoU threshold=50\%. Results are reported on three settings: ``full'' loads the complete checkpoint,  ``enc-only'' loads the Transformer encoder weights, while ``dec-only'' loads the Transformer decoder weights.}
\label{tab:ablation_t5_ckpt}
\end{table*}

\subsection{Comprehensive experimental results}
Table~\ref{tab:vanilla_dvc_results_mean_std_full} provides a more comprehensive summary of our experimental results in the \vanilla setting.  It is the same experimental setting as Table~\ref{tab:vanilla_dvc_results_mean_std}, but we also report additional performance metrics for the segmentation tasks, as well as performance for the \timeanchorShort input formulation.
Table~\ref{tab:vanilla_dvc_results_median} is again under the same experimental setting, but reports median instead of (mean, std) to summarize the 3 repeats for each setting, so that the metrics are less affected by occasional outliers.
\begin{table*}[tbp]
\begin{adjustbox}{width=\linewidth,center}
\begin{tabular}{r | l | l |  r r r r | r r r r r r}
\cmidrule[\heavyrulewidth]{1-13}
\multirow{2}{*}{\textbf{*}} &  \multirowcell{2}{\textbf{Input}\\\textbf{Formulation}}    & \multirow{2}{*}{\textbf{Checkpoint?}}  &    \multicolumn{4}{c}{\textbf{\segonly}}  &   \multicolumn{6}{|c}{\textbf{\segcap}} \\
\cmidrule{4-13}
  &   &    &  mIoU & Precision  & Recall    &   F1  &  mIoU &   F1  &   B@4    &   METEOR  &   CIDEr   &   ROUGE-L \\ 
\cmidrule[\heavyrulewidth]{1-13}

\multicolumn{2}{l}{\textit{YouCook2}} \\ \cmidrule{1-13}
\texttt{0}   & \multicolumn{2}{c|}{Random Segmentation}    & 	20.61	$\pm$	1.04 & 	11.25	$\pm$	0.61 & 	12.49	$\pm$	0.36 & 	10.49	$\pm$	0.59 &   -   &   -   &   -      &   -      &   -      &   -   \\
\cmidrule{2-13}
\texttt{1}   &\multirow{4}{*}{\notemporal}
    &   -                & 	12.99	$\pm$	1.55 & 	12.24	$\pm$	1.08 & 	8.60	$\pm$	0.90 & 	9.39	$\pm$	0.75 & 	16.45	$\pm$	8.72 & 	11.23	$\pm$	5.16 & 	0.17	$\pm$	0.11 & 	0.66	$\pm$	0.04 & 	0.02	$\pm$	0.01 & 	1.99	$\pm$	0.20 \\
\texttt{2}   &    &   T5         & 	24.14	$\pm$	1.07 & 	14.22	$\pm$	0.16 & 	15.09	$\pm$	0.85 & 	14.10	$\pm$	0.44 & 	24.21	$\pm$	1.64 & 	14.20	$\pm$	1.35 & 	0.88	$\pm$	0.04 & 	1.50	$\pm$	0.12 & 	0.09	$\pm$	0.01 & 	3.34	$\pm$	0.27 \\
\texttt{3}    &    &   WikiHow   & 	22.58	$\pm$	1.09 & 	13.39	$\pm$	0.96 & 	14.57	$\pm$	1.19 & 	13.27	$\pm$	1.00 & 	23.33	$\pm$	0.79 & 	14.22	$\pm$	0.94 & 	0.67	$\pm$	0.15 & 	1.47	$\pm$	0.05 & 	0.08	$\pm$	0.01 & 	3.51	$\pm$	0.13 \\
\texttt{4}    &    &   WikiHow T5   & 	\textbf{27.77}	$\pm$	0.09 & 	\textbf{16.68}	$\pm$	1.04 & 	\textbf{18.43}	$\pm$	0.75 & 	\textbf{16.87}	$\pm$	0.62 & 	\textbf{30.26}	$\pm$	1.24 & 	\textbf{20.24}	$\pm$	1.06 & 	\textbf{2.96}	$\pm$	0.28 & 	\textbf{3.49}	$\pm$	0.30 & 	\textbf{0.25}	$\pm$	0.03 & 	\textbf{7.00}	$\pm$	0.42 \\
\cmidrule{2-13}

\texttt{5}   &\multirow{4}{*}{\withanchor}
    &   -                & 	20.13	$\pm$	2.59 & 	13.68	$\pm$	1.78 & 	12.18	$\pm$	1.88 & 	12.01	$\pm$	1.91 & 	18.41	$\pm$	2.65 & 	9.99	$\pm$	1.55 & 	0.08	$\pm$	0.02 & 	0.44	$\pm$	0.04 & 	0.01	$\pm$	0.00 & 	1.33	$\pm$	0.15 \\
\texttt{6}   &    &   T5         & 	20.29	$\pm$	1.30 & 	12.13	$\pm$	2.52 & 	11.43	$\pm$	0.64 & 	11.09	$\pm$	1.38 & 	22.12	$\pm$	1.29 & 	12.56	$\pm$	0.74 & 	0.88	$\pm$	0.23 & 	1.38	$\pm$	0.22 & 	0.08	$\pm$	0.02 & 	3.07	$\pm$	0.39 \\
\texttt{7}    &    &   WikiHow   & 	19.98	$\pm$	0.55 & 	10.54	$\pm$	1.36 & 	12.11	$\pm$	1.29 & 	10.68	$\pm$	1.32 & 	20.84	$\pm$	1.02 & 	11.82	$\pm$	0.64 & 	0.39	$\pm$	0.05 & 	0.99	$\pm$	0.09 & 	0.05	$\pm$	0.00 & 	2.44	$\pm$	0.18 \\
\texttt{8}    &    &   WikiHow T5   & 	20.98	$\pm$	0.69 & 	11.99	$\pm$	1.07 & 	12.49	$\pm$	0.60 & 	11.86	$\pm$	0.82 & 	20.22	$\pm$	0.70 & 	11.20	$\pm$	0.70 & 	0.38	$\pm$	0.08 & 	0.92	$\pm$	0.05 & 	0.05	$\pm$	0.00 & 	2.27	$\pm$	0.13 \\
\cmidrule{2-13}

\texttt{9}    &\multirow{4}{*}{\withtemporal}
    &   -               & 	18.51	$\pm$	1.95 & 	10.85	$\pm$	0.59 & 	11.42	$\pm$	1.16 & 	10.29	$\pm$	0.58 & 	18.71	$\pm$	0.17 & 	9.80	$\pm$	0.80 & 	0.12	$\pm$	0.07 & 	0.48	$\pm$	0.08 & 	0.02	$\pm$	0.01 & 	1.41	$\pm$	0.22 \\
\texttt{10}    &    &   T5       & 	23.02	$\pm$	1.05 & 	13.52	$\pm$	0.76 & 	14.15	$\pm$	0.94 & 	13.23	$\pm$	0.77 & 	23.96	$\pm$	0.08 & 	15.44	$\pm$	0.67 & 	1.32	$\pm$	0.08 & 	1.91	$\pm$	0.07 & 	0.11	$\pm$	0.01 & 	4.20	$\pm$	0.13 \\
\texttt{11}    &    &   WikiHow      & 	21.68	$\pm$	1.93 & 	13.13	$\pm$	1.42 & 	13.88	$\pm$	1.60 & 	12.83	$\pm$	1.41 & 	21.88	$\pm$	0.86 & 	13.15	$\pm$	0.74 & 	0.69	$\pm$	0.19 & 	1.30	$\pm$	0.07 & 	0.07	$\pm$	0.01 & 	3.06	$\pm$	0.13 \\
\texttt{12}    &    &   WikiHow T5     & 	26.51	$\pm$	0.45 & 	15.61	$\pm$	0.61 & 	17.08	$\pm$	0.58 & 	15.82	$\pm$	0.62 & 	28.70	$\pm$	0.92 & 	18.71	$\pm$	0.94 & 	2.58	$\pm$	0.19 & 	3.23	$\pm$	0.10 & 	0.22	$\pm$	0.01 & 	6.45	$\pm$	0.17 \\

\cmidrule[\heavyrulewidth]{1-13}
\multicolumn{2}{l}{\textit{ViTT}} \\ \cmidrule{1-13}
\texttt{13}   & \multicolumn{2}{c|}{Random Segmentation}    & 	21.90	$\pm$	0.15 & 	12.22	$\pm$	0.09 & 	16.12	$\pm$	0.25 & 	12.48	$\pm$	0.10 &   -   &   -   &   -      &   -      &   -      &   -   \\
\cmidrule{2-13}
\texttt{14}   &\multirow{4}{*}{\notemporal}
    &   -                & 	33.85	$\pm$	0.70 & 	23.54	$\pm$	0.36 & 	24.04	$\pm$	0.40 & 	22.98	$\pm$	0.22 & 	32.69	$\pm$	0.71 & 	22.49	$\pm$	0.36 & 	0.11	$\pm$	0.01 & 	3.76	$\pm$	0.35 & 	0.08	$\pm$	0.01 & 	3.86	$\pm$	0.28 \\
\texttt{15}    &    &   T5       & 	37.89	$\pm$	0.10 & 	28.16	$\pm$	1.18 & 	27.15	$\pm$	0.19 & 	27.15	$\pm$	0.53 & 	38.07	$\pm$	0.65 & 	27.39	$\pm$	0.91 & 	0.57	$\pm$	0.03 & 	5.92	$\pm$	0.37 & 	0.16	$\pm$	0.02 & 	6.59	$\pm$	0.69 \\
\texttt{16}    &    &   WikiHow  & 	38.20	$\pm$	0.27 & 	26.95	$\pm$	0.67 & 	27.71	$\pm$	0.25 & 	26.85	$\pm$	0.41 & 	37.80	$\pm$	0.62 & 	26.74	$\pm$	0.81 & 	0.40	$\pm$	0.07 & 	5.48	$\pm$	0.18 & 	0.14	$\pm$	0.01 & 	6.02	$\pm$	0.34 \\
\texttt{17}    &    &   WikiHow T5  & 	\textbf{41.87}	$\pm$	0.26 & 	\textbf{31.75}	$\pm$	1.94 & 	\textbf{31.74}	$\pm$	0.34 & 	\textbf{31.26}	$\pm$	1.10 & 	42.40	$\pm$	0.30 & 	32.01	$\pm$	0.50 & 	\textbf{1.29}	$\pm$	0.07 & 	\textbf{8.10}	$\pm$	0.34 & 	\textbf{0.25}	$\pm$	0.01 & 	\textbf{9.26}	$\pm$	0.39 \\
\cmidrule{2-13}

\texttt{18}   &\multirow{4}{*}{\withanchor}
    &   -                & 	32.19	$\pm$	1.17 & 	20.05	$\pm$	1.89 & 	21.62	$\pm$	0.82 & 	20.04	$\pm$	0.48 & 	32.03	$\pm$	0.14 & 	20.89	$\pm$	0.28 & 	0.05	$\pm$	0.00 & 	2.96	$\pm$	0.13 & 	0.06	$\pm$	0.00 & 	2.93	$\pm$	0.07 \\
\texttt{19}   &    &   T5         & 	34.94	$\pm$	0.37 & 	21.24	$\pm$	0.11 & 	23.95	$\pm$	0.41 & 	22.07	$\pm$	0.21 & 	37.56	$\pm$	0.78 & 	27.50	$\pm$	0.69 & 	0.59	$\pm$	0.09 & 	5.11	$\pm$	0.52 & 	0.16	$\pm$	0.01 & 	6.26	$\pm$	0.56 \\
\texttt{20}    &    &   WikiHow   & 	33.00	$\pm$	0.10 & 	19.13	$\pm$	0.87 & 	22.02	$\pm$	0.13 & 	20.05	$\pm$	0.54 & 	35.14	$\pm$	0.99 & 	22.88	$\pm$	0.41 & 	0.23	$\pm$	0.04 & 	3.51	$\pm$	0.14 & 	0.09	$\pm$	0.01 & 	4.12	$\pm$	0.37 \\
\texttt{21}    &    &   WikiHow T5   & 	34.23	$\pm$	0.55 & 	21.01	$\pm$	1.34 & 	23.26	$\pm$	0.51 & 	21.62	$\pm$	0.94 & 	33.20	$\pm$	1.65 & 	19.63	$\pm$	0.98 & 	0.16	$\pm$	0.02 & 	3.01	$\pm$	0.22 & 	0.08	$\pm$	0.01 & 	3.40	$\pm$	0.36 \\
\cmidrule{2-13}

\texttt{22}    &\multirow{4}{*}{\withtemporal}
    &   -                & 	33.89	$\pm$	0.21 & 	20.75	$\pm$	2.37 & 	23.69	$\pm$	0.08 & 	21.27	$\pm$	1.49 & 	35.37	$\pm$	3.18 & 	22.28	$\pm$	0.49 & 	0.04	$\pm$	0.03 & 	3.42	$\pm$	0.61 & 	0.07	$\pm$	0.01 & 	3.28	$\pm$	0.83 \\
\texttt{23}    &    &   T5      & 	37.78	$\pm$	0.15 & 	25.98	$\pm$	0.20 & 	27.12	$\pm$	0.16 & 	26.05	$\pm$	0.16 & 	38.50	$\pm$	0.55 & 	27.95	$\pm$	0.46 & 	0.75	$\pm$	0.10 & 	6.37	$\pm$	0.39 & 	0.18	$\pm$	0.01 & 	7.19	$\pm$	0.48 \\
\texttt{24}    &    &   WikiHow  & 	37.27	$\pm$	0.08 & 	25.96	$\pm$	0.38 & 	26.87	$\pm$	0.04 & 	25.91	$\pm$	0.21 & 	36.97	$\pm$	0.48 & 	26.37	$\pm$	0.36 & 	0.38	$\pm$	0.06 & 	5.31	$\pm$	0.06 & 	0.13	$\pm$	0.01 & 	5.82	$\pm$	0.23 \\
\texttt{25}    &    &   WikiHow T5  & 	41.64	$\pm$	0.12 & 	31.07	$\pm$	0.67 & 	31.53	$\pm$	0.12 & 	30.84	$\pm$	0.33 & 	\textbf{43.22}	$\pm$	0.72 & 	\textbf{32.49}	$\pm$	\textbf{0.25} & 	1.22	$\pm$	0.08 & 	8.05	$\pm$	0.20 & 	\textbf{0.25}	$\pm$	0.01 & 	9.18	$\pm$	0.45 \\

\cmidrule[\heavyrulewidth]{1-13}
\end{tabular}
\end{adjustbox}
\caption{Dense video captioning performance on YouCook2 and ViTT test sets with the \offsetbased and the \timestampbased formulations. We report the evaluation results (mean $\pm$ std) with models initialized from random weights, T5 checkpoints, WikiHow checkpoints, and T5 checkpoints further pretrained on WikiHow. }
\label{tab:vanilla_dvc_results_mean_std_full}
\end{table*}

\newcommand{\fwidth}{2.1cm}


\begin{table*}[h]
\begin{adjustbox}{width=\linewidth,center}
\begin{tabular}{r|l | l | l |  r r r r | r r r r r r}
\cmidrule[\heavyrulewidth]{1-14}
\multirow{2}{*}{\textbf{*}} & \multirow{2}{*}{\textbf{Dataset}}   &  \multirowcell{2}{\textbf{Input}\\\textbf{Formulation}}    & \multirow{2}{*}{\textbf{Checkpoint}}  &    \multicolumn{4}{c}{\textbf{\segonly}}  &   \multicolumn{6}{|c}{\textbf{\segcap}} \\
\cmidrule{5-14}
    &  &   &    &  mIoU & Precision  & Recall    &   F1  &  mIoU &   F1  &   B@4    &   METEOR  &   CIDEr   &   ROUGE-L \\ 
\cmidrule[\heavyrulewidth]{1-14}

\texttt{0}   & \multirow{9}{*}{YouCook2}   	
    & \multicolumn{2}{c|}{Random Segmentation}    & 	20.10 & 	11.14 & 	12.53 & 	10.69 &   -   &   -   &   -      &   -      &   -      &   -   \\
\cmidrule{3-14}
\texttt{1}   & &\multirow{4}{\fwidth}{\notemporal}
    &   -                & 	12.83 & 	12.80 & 	8.62 & 	9.07 	& 	11.47 & 	8.78 & 	0.22 & 	0.64 & 	0.03 & 	1.95 \\
\texttt{2}   & &    &   T5         & 	24.18 & 	14.19 & 	14.83 & 	13.89 	& 	25.13 & 	14.09 & 	0.86 & 	1.47 & 	0.09 & 	3.39 \\
\texttt{3}    & &    &   WikiHow   & 	22.36 & 	13.11 & 	14.42 & 	12.99 	& 	23.00 & 	14.16 & 	0.66 & 	1.50 & 	0.08 & 	3.48 \\
\texttt{4}    & &    &   WikiHow T5   & 	\textbf{27.81} & 	\textbf{17.21} & 	\textbf{18.16} & 	\textbf{17.01} & 	\textbf{30.97} & 	\textbf{20.57} & 	\textbf{2.85} & 	\textbf{3.48} & 	\textbf{0.24} & 	\textbf{7.02} \\
\cmidrule{3-14}

\texttt{5}    & &\multirow{4}{\fwidth}{\withanchor}
    &   -               & 	18.82 & 	14.42 & 	11.28 & 	11.38 & 	17.18 & 	9.53 & 	0.06 & 	0.45 & 	0.01 & 	1.34 \\
\texttt{6}    & &    &   T5              & 	20.91 & 	13.21 & 	11.48 & 	11.65 & 	22.75 & 	12.87 & 	0.90 & 	1.42 & 	0.08 & 	3.19 \\
\texttt{7}    & &    &   WikiHow      & 	19.76 & 	10.17 & 	11.52 & 	10.19 & 	21.09 & 	11.79 & 	0.37 & 	0.94 & 	0.05 & 	2.35 \\
\texttt{8}    & &    &   WikiHow T5     & 	21.26 & 	12.34 & 	12.83 & 	12.24 & 	20.56 & 	11.35 & 	0.38 & 	0.89 & 	0.05 & 	2.31 \\
\cmidrule{3-14}

\texttt{9}    & &\multirow{4}{\fwidth}{\withtemporal}
    &   -               & 	19.52 & 	10.99 & 	11.70 & 	10.26 	& 	18.77 & 	9.83 & 	0.11 & 	0.52 & 	0.02 & 	1.46 \\
\texttt{10}    & &    &   T5              & 	22.93 & 	13.84 & 	13.78 & 	13.30 	& 	24.00 & 	15.70 & 	1.34 & 	1.90 & 	0.11 & 	4.24 \\
\texttt{11}    & &    &   WikiHow      & 	21.41 & 	13.11 & 	13.88 & 	12.76 	& 	22.08 & 	13.18 & 	0.79 & 	1.30 & 	0.07 & 	3.09 \\
\texttt{12}    & &    &   WikiHow T5     & 	26.61 & 	15.86 & 	17.28 & 	16.08 	& 	28.80 & 	18.41 & 	2.67 & 	3.18 & 	0.23 & 	6.41 \\

\cmidrule[\heavyrulewidth]{1-14}

\texttt{13}   &\multirow{9}{*}{ViTT}
    & \multicolumn{2}{c|}{Random Segmentation}    & 	21.93 & 	12.22 & 	16.07 & 	12.41 &   -   &   -   &   -      &   -      &   -      &   -   \\
\cmidrule{3-14}
\texttt{14}   & &\multirow{4}{\fwidth}{\notemporal}
    &   -                & 	33.74 & 	23.71 & 	23.95 & 	23.10 & 	33.10 & 	22.59 & 	0.12 & 	3.78 & 	0.08 & 	3.87 \\
\texttt{15}    & &    &   T5       & 	37.90 & 	28.28 & 	27.14 & 	27.13 & 	38.35 & 	27.66 & 	0.57 & 	5.85 & 	0.15 & 	6.36 \\
\texttt{16}    & &    &   WikiHow  & 	38.23 & 	26.82 & 	27.78 & 	26.89 & 	37.75 & 	26.92 & 	0.44 & 	5.58 & 	0.14 & 	6.06 \\
\texttt{17}    & &    &   WikiHow T5  & 	\textbf{41.78} & 	\textbf{31.00} & 	\textbf{31.62} & 	\textbf{30.78} & 	42.25 & 	31.85 & 	\textbf{1.34} & 	7.97 & 	\textbf{0.25} & 	\textbf{9.21} \\
\cmidrule{3-14}

\texttt{18}    & &\multirow{4}{\fwidth}{\withanchor}
    &   -               & 	32.62 & 	19.94 & 	22.02 & 	20.27 & 	32.01 & 	20.90 & 	0.05 & 	2.96 & 	0.06 & 	2.95 \\
\texttt{19}    & &    &   T5              & 	34.83 & 	21.20 & 	23.89 & 	22.08 & 	37.36 & 	27.80 & 	0.57 & 	5.31 & 	0.16 & 	6.46 \\
\texttt{20}    & &    &   WikiHow      & 	33.00 & 	19.12 & 	22.09 & 	20.09 & 	35.54 & 	23.09 & 	0.23 & 	3.43 & 	0.09 & 	4.03 \\
\texttt{21}    & &    &   WikiHow T5     & 	33.92 & 	21.12 & 	23.01 & 	21.52 & 	34.04 & 	19.46 & 	0.16 & 	2.96 & 	0.07 & 	3.23 \\
\cmidrule{3-14}

\texttt{22}    & &\multirow{4}{\fwidth}{\withtemporal}
    &   -                & 	33.79 & 	21.21 & 	23.68 & 	21.61 & 	34.56 & 	22.37 & 	0.05 & 	3.12 & 	0.06 & 	2.92 \\
\texttt{23}    & &    &   T5      & 	37.75 & 	25.97 & 	27.13 & 	26.13 & 	38.44 & 	27.94 & 	0.69 & 	6.18 & 	0.18 & 	7.15 \\
\texttt{24}    & &    &   WikiHow  & 	37.22 & 	25.85 & 	26.86 & 	25.84 & 	37.07 & 	26.39 & 	0.37 & 	5.28 & 	0.13 & 	5.73 \\
\texttt{25}    & &    &   WikiHow T5  & 	41.62 & 	30.76 & 	31.52 & 	30.68 & 	\textbf{43.51} & 	\textbf{32.50} & 	1.19 & 	\textbf{8.05} & 	\textbf{0.25} & 	9.02 \\

\cmidrule[\heavyrulewidth]{1-14}
\end{tabular}
\end{adjustbox}
\caption{Performance on the dense video captioning on YouCook2 and ViTT test set with the \offsetbased and the \timestampbased formulations. We report the evaluation results with models initialized from random weights, T5 checkpoints, WikiHow checkpoints, and T5 checkpoints further pretrained on WikiHow.  We ran 3 sets of repeating experiments for each setting, and report the \textbf{median} value on each metric in this Table.}
\label{tab:vanilla_dvc_results_median}
\end{table*}

\subsection{Qualitative examples}
\label{sec:appendix:showcases}
Here we provide more examples for the segmentation subtask and the captioning subtask.

\begin{figure*}[tbp]
    \centering
    \resizebox{0.65\linewidth}{!}{%
    \includegraphics{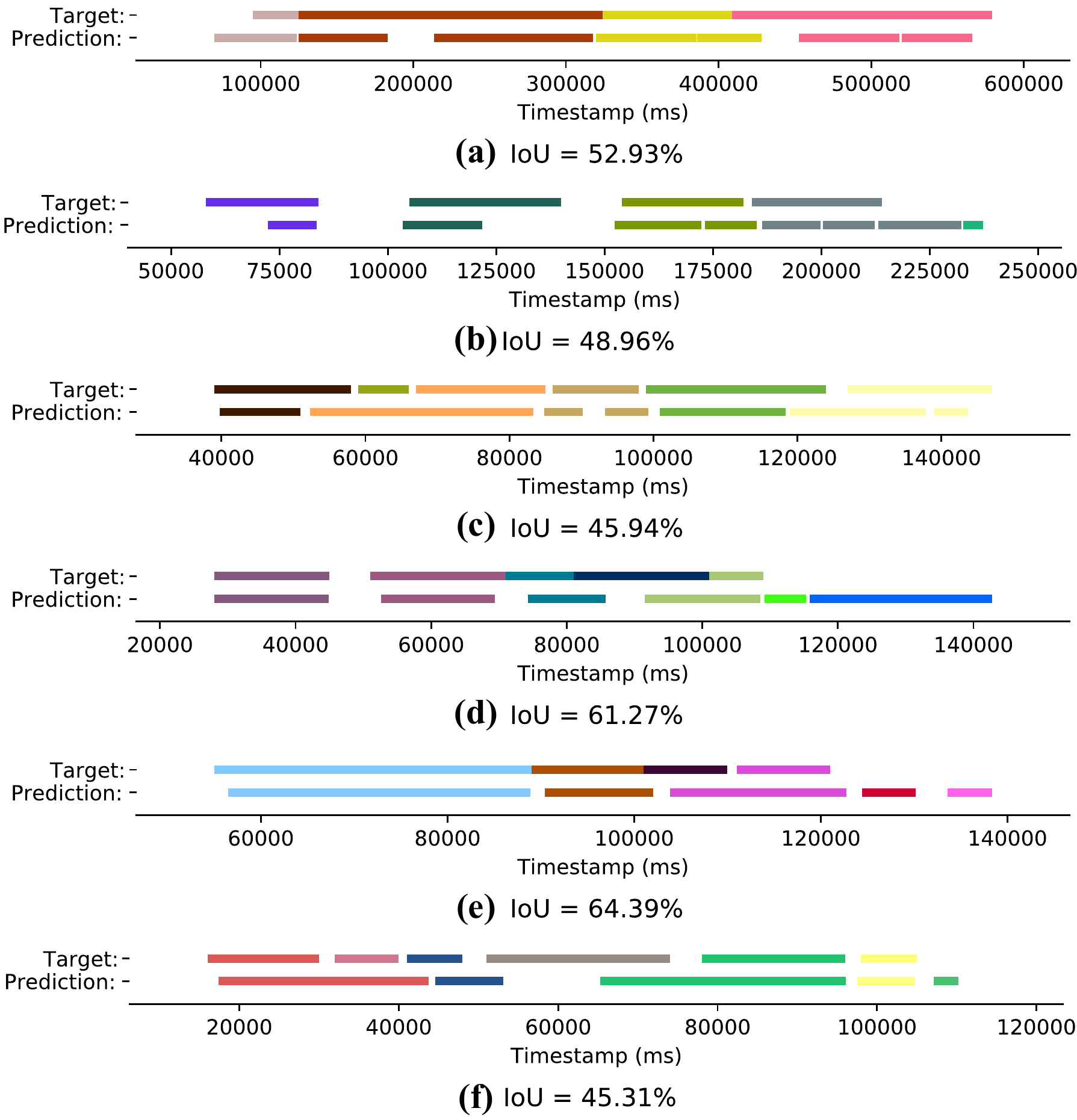}
    }
    \caption{Examples of our \segcap's segmentation performance on YouCook2.
}
    \label{fig:appendix:seg_eg_youcook2}
\end{figure*}

\begin{figure*}[tbp]
    \centering
    \resizebox{0.65\linewidth}{!}{%
    \includegraphics{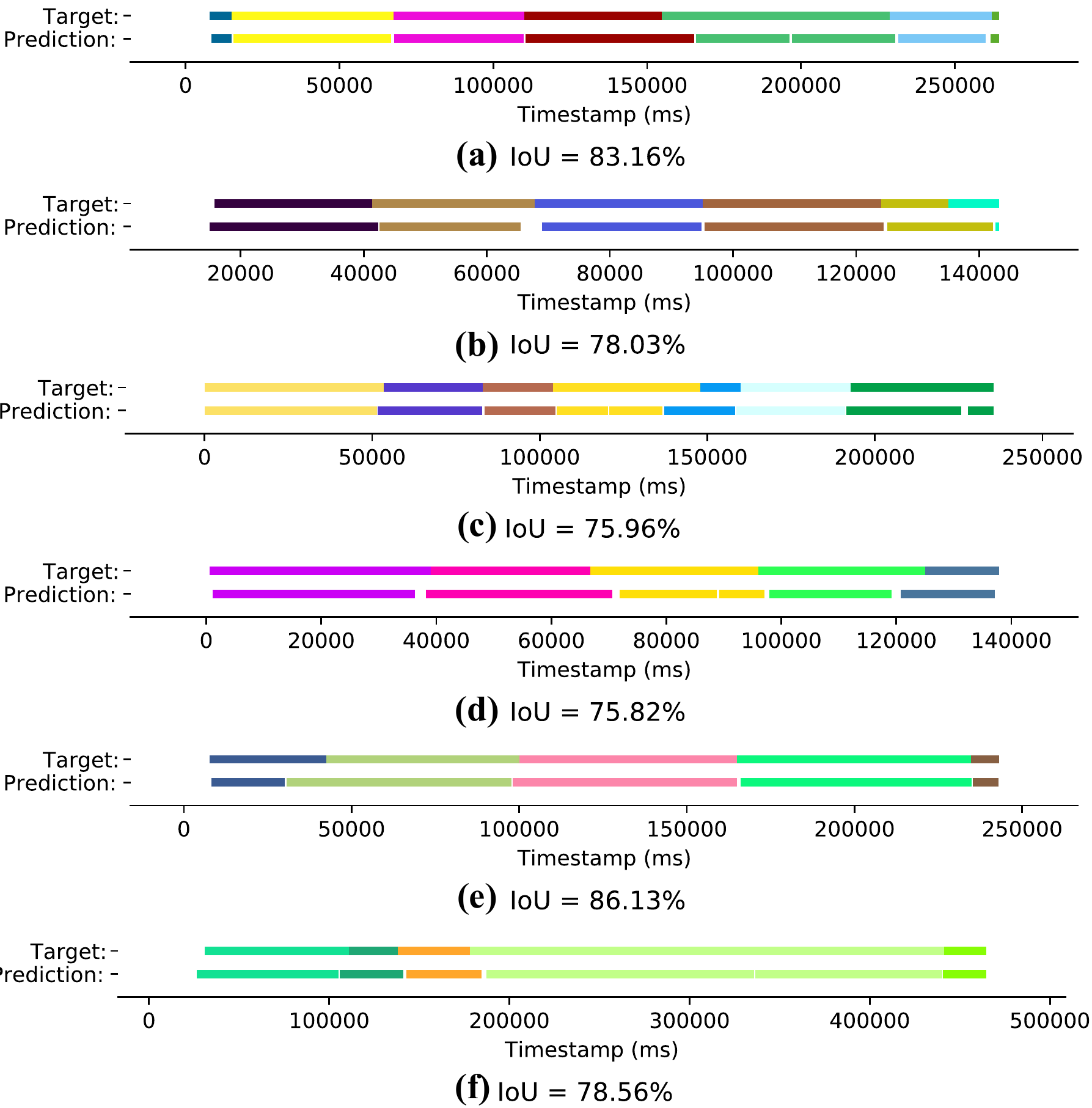}
    }
    \caption{Examples of our \segcap's segmentation performance on ViTT.
}
    \label{fig:appendix:seg_eg_vitt}
\end{figure*}

Figure~\ref{fig:appendix:seg_eg_youcook2} and Figure~\ref{fig:appendix:seg_eg_vitt} illustrate several sets of segmentation results predicted by our \segcap on YouCook2 and ViTT. 
We can see that the predicted segmentation predictions on ViTT (Figure~\ref{fig:appendix:seg_eg_vitt}) are relatively more aligned with the ground-truth. This is because ViTT has a comparably simpler formulation with no gaps between video segments. 
In the more challenging setup on YouCook2 (Figure~\ref{fig:appendix:seg_eg_youcook2}) where the model needs to predict both the start and end point for each segment, the listed examples show that when predictions are mostly correct for a few segments, the IoU scores can be relatively low. 
Common types of segmentation misalignment include: 
\begin{itemize}[noitemsep,topsep=0pt,parsep=0pt,partopsep=0pt]
    \item ``over-segmentation'': the prediction splits a ground-truth span into several sub-chunks (Figure~\ref{fig:appendix:seg_eg_youcook2} (a)(b)(c)); 
    \item ``under-segmentation'': one predicted segment covers several ground-truth events (Figure~\ref{fig:appendix:seg_eg_youcook2} (c)(f)); 
    \item ``prediction-not-covered'': the predicted event is not labeled by the ground-truth annotation (Figure~\ref{fig:appendix:seg_eg_youcook2} (b)(d)(e)).
\end{itemize}

Table~\ref{tab:appendix:caption_example} shows examples of the jointly predicted segments and corresponding captions. We have similar findings as in the main paper that when the segment boundary prediction is well aligned with the ground truth, the corresponding captions are often high-quality as well.

\begin{table*}[tbp]
\begin{adjustbox}{width=\linewidth,center}
\begin{tabular}{c l  r r l}
\cmidrule[\heavyrulewidth]{1-5}
\textbf{Dataset}    &   & \textbf{IoU}  & \textbf{Segment Border (ms)}  & \textbf{Caption} \\ 
\cmidrule[\heavyrulewidth]{1-5}
\multirow{8}{*}{\textit{Youcook2}}
    &   Tgt.    &	\multirow{2}{*}{82.2\%}  &   [49000.0, 67000.0]	&   chop 2 garlic cloves grate ginger about 2 tsp and green onions finely \\
    &   Pred.  &    &	[47114.0, 65354.5] &	chop some garlic ginger and green onions and put them in a bowl \\
\cmidrule{2-5}
    &   Tgt.  &   \multirow{2}{*}{82.4\%}  &   	[73000.0, 117000.0]	&   mix an egg milk and the mashed potatoes \\
    &   Pred. &     &	[72194.0, 109904.5]	&   mix the egg and milk with the potato \\
\cmidrule{2-5}
    &   Tgt.  &   \multirow{2}{*}{81.1\%}  &   	[89000.0, 101000.0]	&   mix and boil the ingredients \\
    &   Pred. &     & [90424.5, 102034.0]    &	add miso paste soy sauce diced vegetables and mushrooms to boiling water \\
\cmidrule{2-5}
    &   Tgt.  &   \multirow{2}{*}{82.1\%}  &   	[18000.0, 48000.0]  &	heat butter in a pan and cook bacon in it \\
    &   Pred. &     & [18224.0, 54254.0]    &	fry pancetta in a pan with bacon \\
\cmidrule[\heavyrulewidth]{1-5}

\multirow{8}{*}{\textit{ViTT}} 
    &   Tgt.  &   \multirow{2}{*}{90.5\%}  &  [147890.0, 189680.0]	 	&   Dipping sticks then cake balls \\
    &   Pred. &                     & [148905.0, 192934.0]	    &   Dipping cake balls in candy melts	 \\
\cmidrule{2-5}
    &   Tgt.  &   \multirow{2}{*}{90.2\%}  &   	[209050.0, 253460.0]	& Buttering in between baking, baking continues   \\
    &   Pred. &                     &  [211445.0, 255634.5]	   &    Brushing the dough with butter	 \\
\cmidrule{2-5}
    &   Tgt.  &   \multirow{2}{*}{90.6\%}  &   [209630.0, 272000.0]		&  Stretching the hamstrings  \\
    &   Pred. &                     &  [213230.0, 269750.0]	   &	 Performing the hamstring stretch \\
\cmidrule{2-5}
    &   Tgt.  &   \multirow{2}{*}{92.3\%}  &   [104000.0, 144000.0]			&  Adding more layers  \\
    &   Pred. &                     &   [105404.0, 145815.0]	  &	 Repeating the same process \\
    
\cmidrule[\heavyrulewidth]{1-5}
\end{tabular}
\end{adjustbox}
\caption{Examples of the jointly predicted segments and corresponding captions for YouCook2 and ViTT generated by our \segcap. Tgt.: Target. Pred.: Prediction.
}
\label{tab:appendix:caption_example}
\end{table*}

\end{document}